\documentclass{article}
\usepackage{iclr2027_conference,times}

\usepackage[table]{xcolor}
\usepackage[ruled,vlined,linesnumbered]{algorithm2e}
\usepackage{hyperref}
\usepackage{fontawesome5}
\usepackage{url}
\usepackage{amssymb}
\usepackage{amsthm}
\usepackage{enumitem}
\usepackage{graphicx}
\usepackage{booktabs}
\usepackage{multirow}
\usepackage{float}

\newcommand{\ours}{\texttt{StepX-OPD}}
\newcommand{\Unif}{\operatorname{Unif}}
\newcommand{\dd}{\mathrm{d}}

\usepackage{amsmath,amsfonts,bm}

\def\eqref#1{equation~\ref{#1}}

\def\1{\bm{1}}

\DeclareMathAlphabet{\mathsfit}{\encodingdefault}{\sfdefault}{m}{sl}
\SetMathAlphabet{\mathsfit}{bold}{\encodingdefault}{\sfdefault}{bx}{n}

\newcommand{\E}{\mathbb{E}}

\newcommand{\KL}{D_{\mathrm{KL}}}

\usepackage{wrapfig}
\usepackage{titletoc}

\newtheorem{proposition}{Proposition}[section]
\newtheorem{theorem}[proposition]{Theorem}
\newtheorem{lemma}[proposition]{Lemma}

\title{\ours: Rethinking Multi-Teacher \\ On-Policy Distillation in Diffusion Models}

\author{\bfseries Chenru Wang$^{1*}$, Xuanyang Zhang$^{1\dagger}$, Yucheng Han$^{1}$, Wenbo Nie$^{1*}$, Wei Cheng$^{1}$, \\[8pt]
\bfseries Xianfang Zeng$^{1}$, Gang YU$^{1}$, Chi Zhang$^{2\ddagger}$ \\[6pt]
\normalfont $^{1}$StepFun \quad $^{2}$Westlake University \\[3pt]
\multicolumn{1}{@{}l@{}}{\makebox[\textwidth][c]{\raisebox{-13.5pt}[\height][\depth]{\normalfont \href{https://github.com/WinterCCC/StepX-OPD}{\faGithub~\textcolor{blue!70!black}{GitHub}} \quad $\vert$ \quad \href{https://huggingface.co/ShimaRinC/StepX-OPD-SD3.5}{\raisebox{-0.12em}{\includegraphics[height=0.95em]{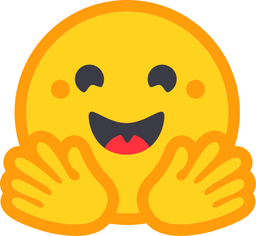}}~\textcolor{blue!70!black}{Hugging Face}} \quad $\vert$ \quad \href{https://winterccc.github.io/Step-OPD/}{\faGlobe~\textcolor{blue!70!black}{Project Page}} \quad $\vert$ \quad \href{https://huggingface.co/spaces/hugging-apps/stepx-opd}{\faPlayCircle~\textcolor{blue!70!black}{Demo}}}}}}

\iclrfinalcopy
\begin{document}
\raggedbottom
\setlength{\textfloatsep}{20pt plus 0pt minus 4pt}
\setlength{\intextsep}{12pt plus 0pt minus 2pt}
\setlength{\floatsep}{12pt plus 0pt minus 2pt}

\maketitle
\lhead{}
\begingroup
\renewcommand{\thefootnote}{\fnsymbol{footnote}}
\footnotetext[1]{Work done during an internship at StepFun.}
\footnotetext[2]{Project lead.}
\footnotetext[3]{Corresponding author.}
\endgroup

\begin{abstract}A deployable text-to-image diffusion model must hold several capabilities at once, such as instruction following, text rendering and human preference, and must sample in a few steps. Capabilities and few-step sampling are built in separate stages: a full-step model first consolidates capabilities through multi-teacher on-policy distillation, then serves as a fixed teacher for step distillation. Every full-step model we evaluate scores lower once compressed, and the second stage cannot begin until the first finishes. We introduce \ours{}, which merges the two stages: one student acquires every capability while learning to sample in a few steps, so frozen specialists supervise it on the states its own rollout visits rather than on a longer trajectory it never runs. Motivated by a decomposition of the rollout's KL, we also regress the student's velocity on the low-noise part of the rollout. Separately, we find that capability integration makes models write prompt words into images on prompts that never asked for text, and we give a variant that lowers it. We evaluate on SD3.5-Medium and FLUX.1-dev across instruction following, text rendering, image quality and human preference. On both architectures, \ours{} attains the highest normalized average score among the four-step methods compared and exceeds the forty-step Flow-OPD model, and on SD3.5-Medium it trains with about 37\% fewer GPU-hours than sequential DiffusionOPD followed by step distillation.
\end{abstract}

\begin{figure}[H]
\vspace{0pt}
\centering
\includegraphics[trim=0 14 0 16, clip, width=\linewidth]{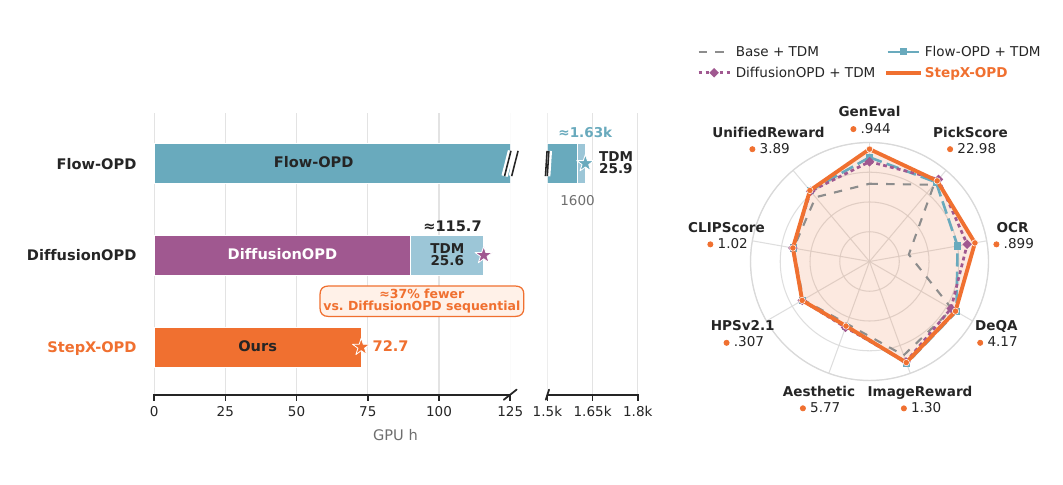}
\caption{Training cost for the displayed optimization stages (left) and four-step performance across nine benchmarks after fixed axis-wise normalization (right). }
\label{fig:teaser}
\end{figure}

\section{Introduction}
\label{sec:intro}

Flow matching and diffusion models are widely used for text-to-image generation \citep{lipman2023flowmatching,esser2024sd3}. A model prepared for deployment must meet two needs at once: it must be capable across composition, text rendering, image quality, and human preference, and it must be fast, measured by how few sampling steps its inference requires. Models meet these needs through two separate lines of post-training, one that integrates capabilities into a single model and one that compresses its sampling trajectory into a few steps. These two lines are developed independently, each advancing its own need while leaving the other untouched. Yet both needs are placed on the same deployed model, so capability integration and step distillation constrain that model together.

Existing methods address these two deployment needs separately. Capability integration builds on on-policy distillation, which supervises the student on states the student itself generates \citep{agarwal2024gkd}. In continuous image generation, DiffusionOPD~\citep{li2026diffusionopd}, Flow-OPD~\citep{fang2026flowopd}, and DanceOPD~\citep{zhou2026danceopd} use analytic transition KL or local regression to integrate capabilities, but retain full-step sampling. Step distillation, by contrast, starts from a teacher whose capabilities are already fixed. DMD~\citep{yin2024dmd} and DMD2\citep{yin2024dmd2} use a frozen teacher score and an auxiliary fake-score estimator, while Qwen-Image-Flash~\citep{wu2026qwenflash} combines a base teacher with a post-trained teacher. TDM~\citep{luo2025tdm}, CDM~\citep{liu2026ctdm}, and SenseFlow~\citep{ge2026senseflow} move supervision toward student-generated trajectories or nearby states, which improves the few-step trajectory. These compression methods still treat teacher capabilities as inputs. A multi-capability few-step model therefore continues to rely on a serial pipeline: multi-teacher capability integration first produces a full-step multi-capability model, and step distillation then treats that model as a fixed teacher for few-step compression. The serial pipeline lengthens training, because compression starts only once the full-step capability model finishes. Fig.~\ref{fig:teaser} reports the training cost each optimization stage carries. Beyond this training cost, it leaves open the question: \textit{do the capabilities integrated into the full-step model still hold once its sampling trajectory is compressed to a few steps?}

We test this directly by comparing each full-step capability model with its few-step compressed counterpart: Tab.~\ref{tab:main_results} reports lower aggregate scores for every one of them after compression, with larger declines on individual metrics. Fig.~\ref{fig:spillover} shows that these full-step OPD models already render prompt words into images more often than the base model, on prompts that neither request nor forbid text, before their trajectories are shortened. Transition matching in continuous-image OPD and marginal matching in DMD~\citep{yin2024dmd} and TDM~\citep{luo2025tdm} both admit reverse-KL interpretations over different distributions and states. Theorem~\ref{thm:theory:decomposition} analyzes this relationship through a reverse-KL decomposition.

Motivated by this decomposition, we propose \ours{}, which integrates capabilities and compresses sampling in one training run, without an intermediate full-step OPD model. Frozen specialists supervise the student directly on the states its current few-step rollout visits, with each mini-batch going to the specialist for its training source. The same decomposition motivates \ours{} to also regress the student's velocity on the low-noise part of the rollout, telling it not only which distribution to land in but also which direction to move.

We evaluate \ours{} on SD3.5-Medium~\citep{esser2024sd3} and FLUX.1-dev~\citep{labs2025flux1kontextflowmatching} across GenEval~\citep{ghosh2023geneval}, OCR accuracy~\citep{chen2023textdiffuser,cui2025paddleocr30technicalreport}, DeQA~\citep{you2025deqa}, PickScore~\citep{kirstain2023pickscore} and their normalized Average, against sequential Flow-OPD + TDM and DiffusionOPD + TDM baselines. On both architectures, \ours{} achieves the highest Average among the few-step systems, 0.8901 and 0.8902, and exceeds the forty-step Flow-OPD model, and on SD3.5-Medium it trains with about 37\% fewer GPU-hours than DiffusionOPD + TDM. On prompts that do not request text, recurrence rises from 21 images for the SD3.5-Medium base to 218 for DiffusionOPD, and \ours{}, trained with the DiffusionOPD teachers, reduces it to 94, and to 84 with its precise-teacher variant.

Our contributions are summarized as follows:
\begin{itemize}[leftmargin=4mm]
    \item We introduce \ours{}, which trains capability integration and few-step compression together on the student's current rollout, with the routed specialist supplying both the distribution-matching and the velocity-regression targets.
    \item We measure unintended prompt-word rendering on prompts that neither request nor forbid text, find it rises with capability integration, and give a variant that reduces it.
    \item We show under a common protocol that every full-step capability model we evaluate scores lower once compressed, and that \ours{} attains the highest Average among the few-step systems on both architectures and, on SD3.5-Medium, trains with about 37\% fewer GPU-hours than sequential DiffusionOPD + TDM.
\end{itemize}

\section{Related work}
\label{sec:related}

\paragraph{Step distillation.}
Within distribution-matching methods, DMD uses the difference between real and fake score fields as a reverse-KL gradient. It also adds regression on teacher-generated noise--image pairs, so part of the update still depends on a multi-step teacher sampler \citep{yin2024dmd}. DMD2 removes this paired regression, uses two-time-scale updates to stabilize fake-score learning, adds adversarial training, and uses fixed-schedule backward simulation for multi-step students \citep{yin2024dmd2}. Decoupled DMD separates CFG augmentation from distribution matching and gives them different re-noising schedules, while retaining DMD2's fixed simulation and fake-score training \citep{liu2026decoupleddmd}. A data-dependent alternative, Moment Matching, matches conditional clean-data means. SiD derives data-free score identities \citep{salimans2024momentmatching,zhou2024sid}. TDM instead matches intermediate trajectory marginals over step-dependent intervals, without instance-level trajectory matching \citep{luo2025tdm}. SenseFlow adapts DMD to large flow backbones through implicit alignment, intra-segment guidance, real-data supervision, and a VFM discriminator, while keeping coarse fixed anchors \citep{ge2026senseflow}. CDM replaces fixed anchors with dynamically sampled on- and off-trajectory states. This broadens supervision but adds training-time evaluations, while inference uses the same number of steps \citep{liu2026ctdm}. Qwen-Image-Flash blends base and reinforcement-trained teacher scores with a fixed step schedule. The selected blend is more stable than its specialist-only setting \citep{wu2026qwenflash}. TDM-R1 \citep{luo2026tdmr1} reinforces a few-step TDM generator directly with non-differentiable rewards by learning per-step surrogate rewards along its deterministic trajectory, so its capabilities come from reward signals on the few-step generator itself, and existing specialist models do not serve as teachers. The distillation methods above still provide fixed capabilities to the distillation objective; Qwen's blend is also fixed in advance rather than chosen from the states the student visits, motivating the on-policy routing methods discussed next.

\paragraph{On-policy distillation in diffusion models.}
On-policy distillation lets a teacher supervise the states the student visits during generation, rather than states drawn from an offline trajectory or dataset. In multimodal language models, Vision-OPD \citep{yuan2026visionopd} applies it as self-distillation: the same model acts as the teacher when it sees evidence-centered crops and as the student when it sees the full image, and training matches their next-token distributions along the student's rollouts. In diffusion models, DiffusionOPD \citep{li2026diffusionopd} lifts this idea from the discrete token transitions used in language models to the continuous-state denoising process, modeling each step as a Gaussian transition and deriving a closed-form per-step KL objective in place of a PPO-style surrogate. Flow-OPD \citep{fang2026flowopd} first trains a domain-specialized teacher per reward with single-reward GRPO, then merges the teachers into a cold-started student via on-policy sampling and task-routing labeling, adding a Manifold Anchor Regularization term in which a task-agnostic teacher anchors generation to a high-quality manifold. DanceOPD \citep{zhou2026danceopd} treats each frozen capability as a velocity field over the denoising state space, routing each sample to one field, querying it once on a low-noise state from the student's rollout, and training with a plain velocity mean-squared-error objective. A second arrangement applies on-policy supervision after step distillation rather than before it. D-OPSD \citep{jiang2026dopsd} tunes an already step-distilled few-step model by having the same network act as teacher and student under different conditioning, the student seeing only the text prompt while the teacher also sees the target image, so it learns new concepts or styles on its rollout without degrading its distilled few-step behavior. Both arrangements treat capability integration and step-count reduction as separate stages; \ours{} places multi-teacher on-policy supervision inside the few-step distillation objective, carrying out both in a single stage.

For a broader review of unintended text rendering, see App.~\ref{sec:app:text-rendering-related}.

\section{Preliminaries}
\label{sec:prelim}
\setlength{\abovedisplayskip}{2pt}
\setlength{\belowdisplayskip}{2pt}
\setlength{\jot}{1pt}

\noindent\textbf{Flow matching.} We use the rectified-flow convention $t\in[0,1]$, where $t=1$ is pure noise and $t=0$ is clean \citep{lipman2023flowmatching,esser2024sd3}, and denote the condition by $c$. The linear path is
\begin{equation}
z_t=(1-t)x_0+t\epsilon .
\label{eq:flow_interp}
\end{equation}
Here, $x_0$ is a clean sample and $\epsilon\sim\mathcal N(0,I)$, and differentiating Eq.~\ref{eq:flow_interp} with respect to $t$ gives the target velocity $\mathrm dz_t/\mathrm dt=\epsilon-x_0$ that the student velocity $v_\theta$ regresses onto. We write $v_\phi$ for the teacher velocity, trained the same way, and use $1=\sigma_0>\cdots>\sigma_N=0$ for the student grid. Sampling advances along that grid by the deterministic Euler update
\begin{equation}
z_{\sigma_{i+1}}=z_{\sigma_i}+(\sigma_{i+1}-\sigma_i)\,v_\theta(z_{\sigma_i},\sigma_i),
\label{eq:euler_step}
\end{equation}
whose step $\sigma_{i+1}-\sigma_i$ is negative because the grid decreases from noise to clean.

\noindent\textbf{Distribution matching.} At the clean terminal distribution, DMD~\citep{yin2024dmd,yin2024dmd2} minimizes the reverse KL $\mathrm D_{\mathrm{KL}}(q_\theta^{\mathrm{clean}}\,\|\,p_\phi^{\mathrm{clean}})$. TDM \citep{luo2025tdm} extends the same marginal matching to intermediate rollout distributions:
\begin{equation}
\mathcal L_{\mathrm{TDM}}=\sum_{i=1}^{N}\mathbb E_{\tau\sim\pi_i}\!\left[\lambda_\tau\,\mathrm D_{\mathrm{KL}}(q_{\theta,\tau}^i\,\|\,p_{\phi,\tau})\right].
\label{eq:tdm_objective}
\end{equation}
Here, $q_\theta^i=\operatorname{Law}(z_{\sigma_i})$, $q_{\theta,\tau}^i=k_{\tau|\sigma_i}q_\theta^i$, and $\operatorname{supp}(\pi_i)\subset[\sigma_i,\sigma_{i-1}]$; $p_{\phi,\tau}$ is the teacher marginal and $\lambda_\tau$ is the timestep weight. Its gradient is estimated by
\begin{equation}
\nabla_\theta\mathcal L_{\mathrm{TDM}}\approx\sum_{i=1}^{N}\mathbb E_{\tau\sim\pi_i,\,z_\tau\sim q_{\theta,\tau}^i}\!\left[\lambda_\tau J_\tau^\top(s_\psi-s_{\phi,\tau})\right].
\label{eq:tdm_grad}
\end{equation}
Here, $s_\psi\approx s_{\theta,\tau}^i=\nabla_{z_\tau}\log q_{\theta,\tau}^i$, $s_{\phi,\tau}=\nabla_{z_\tau}\log p_{\phi,\tau}$, and $J_\tau=\partial z_\tau/\partial\theta$.

\noindent\textbf{On-policy distillation.} OPD \citep{li2026diffusionopd,fang2026flowopd} matches the conditional transition reverse KL at the same student-visited state:
\begin{equation}
\mathrm D_{\mathrm{KL}}\!\left(q_{\theta,i}(\cdot|z_{\sigma_i})\,\|\,p_{\phi,i}(\cdot|z_{\sigma_i})\right)=\omega_i\left\|v_\theta(z_{\sigma_i},\sigma_i)-v_\phi(z_{\sigma_i},\sigma_i)\right\|^2.
\label{eq:opd_kl}
\end{equation}
Here, $q_{\theta,i}(\cdot|z_{\sigma_i})$ and $p_{\phi,i}(\cdot|z_{\sigma_i})$ are the student and teacher kernels, $\Delta_i=\sigma_i-\sigma_{i+1}>0$, and $g_i$ is the diffusion coefficient at $\sigma_i$; they share covariance $g_i^2\Delta_i I$, and $\omega_i=\frac{\Delta_i}{2}(\frac1{g_i}+\frac{g_i(1-\sigma_i)}{2\sigma_i})^2$ for $0<\sigma_i<1$. The later batch-gated velocity MSE is an unweighted surrogate of this term, and the weight $w_b$ that App.~\ref{sec:app:theory} derives is a separate marginal-KL weight.

\section{Method}
\label{sec:method}

In this section, we introduce our proposed \ours{}, which trains one $N$-step student in a single run. The training source of each mini-batch picks the frozen specialist that supplies the target for that batch. The specialist supervises states built on the student's own few-step rollout, through the distribution-matching loss of TDM~\citep{luo2025tdm} and a velocity regression on the low-noise part of the rollout, which the decomposition in Sec.~\ref{sec:method:vmse} motivates. Alg.~\ref{alg:stepx_opd} in App.~\ref{sec:app:algorithm} summarizes one training step, and the rest of this section describes source routing, the supervision states, and the low-noise velocity regression in turn.

\subsection{Source routing to frozen specialists}
\label{sec:method:routing}

Each training step draws its mini-batch from a single training source $d$ among $\{\mathcal C_d\}$, rather than mixing sources within a batch, and this source choice determines which model serves as the active target $v^{\mathrm{tgt}}$ for every supervision state drawn from that batch. During an early warmup, $v^{\mathrm{tgt}}$ is the frozen initial student $v_{\theta_0}^{\mathrm{stu}}$, letting the student's own few-step rollout settle before any teacher enters. Afterward, a static lookup $r$ maps the batch's source $d$ to one of $M$ frozen specialists $\{v_m^{\mathrm{tea}}\}_{m=1}^M$, and $v^{\mathrm{tgt}}$ becomes the routed specialist $v_{r(d)}^{\mathrm{tea}}$. Because $r$ depends only on the sampled source and is fixed before the forward pass begins, every data-parallel process resolves the same routed teacher for a given step without exchanging any information. Each training source corresponds to one capability a specialist was trained to supply, so routing by source sends every mini-batch to the one specialist whose target behavior matches the capability that source is meant to teach.

\subsection{Supervision states on the randomized rollout}
\label{sec:method:points}

Inspired by CDM~\citep{liu2026ctdm}, DiDi-Instruct~\citep{zheng2026didi}, and the public implementation of TDM~\citep{luo2025tdm}, the student rolls out on a grid $1=\sigma_0>\sigma_1>\dots>\sigma_N=0$ with both ends pinned and interior nodes redrawn before every training step. The rollout produces states only at these nodes. To supervise at a time between two nodes, \ours{} constructs a state there.

Given this randomized rollout grid, \ours{} first selects one segment $[\sigma_\ell,\sigma_u]$ uniformly at random, then draws a base time $\tau_{\text{base}}$ uniformly within it, with $\sigma_\ell\le\tau_{\text{base}}<\sigma_u$. Let $z_{\text{start}}$ be the state at $\sigma_u$ and $z_{\text{end}}$ the state at $\sigma_\ell$, and interpolate between them using
\begin{align}
 x_{\text{base}} &= \gamma\, z_{\text{start}} + (1-\gamma)\, \operatorname{sg}[z_{\text{end}}]. \label{eq:xbase}
\end{align}
where $\gamma = \frac{\tau_{\text{base}}-\sigma_\ell}{\sigma_u-\sigma_\ell}$. Here $\operatorname{sg}$ denotes the stop-gradient. When $\tau_{\text{base}}=\sigma_\ell$, $\gamma=0$ and Eq.~\ref{eq:xbase} reduces to $x_{\text{base}}=\operatorname{sg}[z_{\text{end}}]$, so the construction includes direct re-noising from the lower endpoint as a special case. The two rollout states being interpolated are related by Eq.~\ref{eq:euler_step}, since that is the update the rollout visiting them took, which lets $x_{\text{base}}$ be characterized exactly.

The interpolated state $x_{\text{base}}$ is exactly the state a first-order Euler step from $z_{\text{start}}$ reaches at $\tau_{\text{base}}$, and App.~\ref{sec:app:theory} proves this as Lemma~\ref{lem:interp}.

Conditioned on the grid and $\tau_{\text{base}}$, we sample the supervision time $\tau\sim U[\tau_{\text{base}},\sigma_u]$.
We obtain $z_\tau$ from $x_{\text{base}}$ by the exact forward SDE transition from $\tau_{\text{base}}$ to $\tau$, whose coefficients App.~\ref{sec:app:theory} derives by matching moments with the Euler re-noising law of Lemma~\ref{lem:interp}. With $\alpha_t=1-t$ and $\sigma_t=t$, this re-noising step is
\begin{equation}
z_\tau = \kappa\, x_{\text{base}} + \beta\,\epsilon, \label{eq:renoise}
\end{equation}
where $\epsilon$ is a standard Gaussian draw, $\kappa = \frac{\alpha_\tau}{\alpha_{\tau_{\text{base}}}}$, and
$\beta = \sqrt{\max\!\big(\sigma_\tau^2 - (\kappa\,\sigma_{\tau_{\text{base}}})^2,\ 0\big)}$. Because $\tau$ remains inside the sampled segment $[\tau_{\text{base}},\sigma_u]$, Lemma~\ref{lem:interp} applies directly and $z_\tau$ carries $x_{\text{base}}$'s own conditional law forward to $\tau$ exactly.

\subsection{Low-noise velocity regression}
\label{sec:method:vmse}

Theorem~\ref{thm:theory:decomposition} shows that, under its assumptions,
\begin{equation}
\mathrm D_{\mathrm{KL}}(q_b\|p_b)
= \mathbb E_T\,\mathrm D_{\mathrm{KL}}(q_T\|p_T)
+ \mathbb E\big[w_b(T)\,\|v_q(Z_T,T)-v_p(Z_T,T)\|^2\big],
\label{eq:vmse_decomposition}
\end{equation}
where $q_t$ and $p_t$ are the rollout state's own distribution and the teacher distribution at time $t$, $v_q$ and $v_p$ are the velocity fields induced by these two forward-smoothed laws, and $Z_T$ is the state at a time $T$ drawn uniformly on the segment. The weight is
\begin{equation}
w_b(t) = (t_i-t)\,\frac{1-t}{t}, \qquad b<t<t_i,
\label{eq:vmse_weight}
\end{equation}
where $b$ is the base time and $t_i$ the segment's upper endpoint. Because $w_b$ increases as $t$ decreases toward the low-noise end of the segment, Theorem~\ref{thm:theory:decomposition} motivates adding velocity regression as an auxiliary term on the low-noise part of the rollout. App.~\ref{sec:app:theory} gives the full statement, the proof, and the bound relating the student network to the field its rollout induces. See App.~\ref{sec:theory:gap} for how the implemented objective below relates to the quantities in the theorem.

For each sampled state $z_\tau$, the active target $v^{\mathrm{tgt}}$ selected by source routing (see Sec.~\ref{sec:method:routing}) supplies the instantaneous classifier-free guided velocity $\tilde v^{\mathrm{tgt}}=v_{\mathrm{cfg}}^{\mathrm{tgt}}(z_\tau,\tau,c)=v^{\mathrm{tgt}}(z_\tau,\tau,\varnothing)+w\,[v^{\mathrm{tgt}}(z_\tau,\tau,c)-v^{\mathrm{tgt}}(z_\tau,\tau,\varnothing)]$ with guidance scale $w$.
The trainable student $v_\theta^{\mathrm{stu}}$ makes an additional forward pass at the same state and time and regresses toward this frozen target through
\begin{equation}
\mathcal{L}_{\text{v}}(\tilde v^{\mathrm{tgt}})=\frac{1}{N_{\mathrm{elt}}}\sum_{j=1}^{B}\big\|v_\theta^{\mathrm{stu}}(z_{\tau_j},\tau_j,c)-\operatorname{sg}\big[\tilde v_j^{\mathrm{tgt}}\big]\big\|_2^2,
\label{eq:vmse_loss}
\end{equation}
where $B$ is the number of sampled states in the batch, and $N_{\mathrm{elt}}$ is the number of packed tensor elements those states carry. Eq.~\ref{eq:vmse_loss} is the unweighted surrogate of the conditional-transition KL in Eq.~\ref{eq:opd_kl}: dropping the weight $\omega_i$ and averaging the squared distance elementwise rather than summing it yields the $1/N_{\mathrm{elt}}$ factor, and the target becomes the guided, per-step $\tilde v^{\mathrm{tgt}}$ rather than a single fixed teacher velocity $v_\phi$. The auxiliary term is gated once per mini-batch, using the batch-mean sampled time: with training step $s$, configured start step $s_{\mathrm{v}}$, sampled times $\{\tau_j\}_{j=1}^B$, and threshold $\tau_{\mathrm{v}}$,
\begin{equation}
g_{\mathrm{v}}=\mathbf{1}\!\left\{s\ge s_{\mathrm{v}}\right\}\,\mathbf{1}\!\left\{\bar\tau\le\tau_{\mathrm{v}}\right\},
\qquad \bar\tau=\frac{1}{B}\sum_{j=1}^{B}\tau_j. \label{eq:vmse_gate}
\end{equation}
This supplies direct local velocity supervision near the clean endpoint, where the capability signal must be transferred through the student's on-policy states. The same target $\tilde v^{\mathrm{tgt}}$ also supplies the teacher score for the distribution-matching term $\mathcal{L}_{\mathrm{TDM}}(\tilde v^{\mathrm{tgt}})$, which follows the gradient estimator of Eq.~\ref{eq:tdm_grad} with the teacher score $s_{\phi,\tau}$ evaluated at $\tilde v^{\mathrm{tgt}}$. The total objective is
\begin{equation}
\mathcal{L}_{\mathrm{total}} = \mathcal{L}_{\mathrm{TDM}}(\tilde v^{\mathrm{tgt}}) + w_{\text{v}} g_{\mathrm{v}}\,\mathcal{L}_{\mathrm{v}}(\tilde v^{\mathrm{tgt}}). \label{eq:total_loss}
\end{equation}
See App.~\ref{sec:app:precise} for an optional target that replaces this instantaneous velocity with an average taken along a short guided rollout from the supervision state.

\begin{table*}[t]
\centering
\fontsize{8}{10}\selectfont
\setlength{\tabcolsep}{4pt}
\caption{Comparison of text-to-image scores across methods, at forty and at four sampling steps, for both base architectures evaluated. The lower per-architecture block reports the same models after step distillation.}
\label{tab:main_results}
\begin{tabular}{l|lrrrrrr}
\toprule
Base Model & Model & NFEs & GenEval $\uparrow$ & OCR $\uparrow$ & DeQA $\uparrow$ & PickScore $\uparrow$ & Average $\uparrow$ \\
\midrule
\multirow{11}{*}{SD3.5-Medium}
 & Base               &80 & 0.666 & 0.529 & 4.045 & 21.79 & 0.7105 \\
 & GenEval Teacher    & 80 & 0.946 & 0.597 & 3.965 & 21.66 & 0.7923 \\
 & OCR Teacher        & 80 & 0.656 & 0.924 & 3.849 & 21.66 & 0.7957 \\
 & PickScore Teacher  & 80 & 0.744 & 0.629 & 4.166 & 23.30 & 0.7756 \\
 & Flow-OPD           & 80 & 0.905 & 0.885 & 4.268 & 22.66 & 0.8788 \\
 & DiffusionOPD       & 80 & 0.945 & 0.930 & 4.113 & 23.68 & 0.9022 \\

\cmidrule(lr){2-8}
 & Base + TDM         & 4  & 0.653 & 0.335 & 4.002 & 21.87 & 0.6573 \\
 & Flow-OPD + TDM     & 4  & 0.875 & 0.751 & 4.217 & 22.60 & 0.8345 \\
 & DiffusionOPD + TDM & 4  & 0.838 & 0.833 & 3.952 & 23.41 & 0.8403 \\
 \rowcolor{cyan!8}
 \cellcolor{white} & \ours{}~(Ours)            & 4  & 0.944 & 0.899 & 4.170 & 22.98 & 0.8901 \\
\midrule
\multirow{10}{*}{FLUX.1-dev}
 & Base                 & 40 & 0.655 & 0.588 & 4.298 & 21.96 & 0.7368 \\
 & GenEval Teacher   & 40 & 0.900 & 0.663 & 4.333 & 22.06 & 0.8195 \\
 & OCR Teacher       & 40 & 0.670 & 0.942 & 4.254 & 21.99 & 0.8270 \\
 & PickScore Teacher & 40 & 0.690 & 0.678 & 4.153 & 23.57 & 0.7765 \\
 & Flow-OPD       & 40 & 0.880 & 0.913 & 4.247 & 23.50 & 0.8864 \\
 & DiffusionOPD   & 40 & 0.892 & 0.933 & 4.225 & 23.58 & 0.8943 \\
\cmidrule(lr){2-8}
 & Base + TDM           & 4  & 0.632 & 0.557 & 4.340 & 21.94 & 0.7252 \\
 & Flow-OPD + TDM       & 4  & 0.864 & 0.901 & 4.087 & 23.30 & 0.8695 \\
 & DiffusionOPD + TDM   & 4  & 0.882 & 0.925 & 4.005 & 23.32 & 0.8761 \\
 \rowcolor{cyan!8}
 \cellcolor{white} & \ours{}~(Ours)               & 4  & 0.890   & 0.929   & 4.233   & 23.28   & 0.8902    \\
\bottomrule
\end{tabular}
\vspace{0pt}
\end{table*}

\section{Experiments}
\label{sec:exp}

In this section, we present the experiments that evaluate the effectiveness of \ours{}. See App.~\ref{sec:app:additional} for additional experiments.

\subsection{Experimental setup}
\label{sec:exp:setup}

\noindent\textbf{Model.} We evaluate two base architectures, SD3.5-Medium \citep{esser2024sd3} and FLUX.1-dev~\citep{labs2025flux1kontextflowmatching}. For each, the joint OPD models and task-specialist teachers use forty-step generation and the step-distilled students, including \ours{}, use four steps. Every evaluated model is sampled at $512\times512$ resolution. On SD3.5-Medium, full-step models use a classifier-free guidance scale of 4.5 and four-step students a scale of 1.0. FLUX.1-dev instead carries a distilled guidance embedding, set to 3.5 at forty steps and 1.0 at four steps, with classifier-free guidance disabled throughout.

\noindent\textbf{Benchmark.} We evaluate each base model, its task specialists, and its joint OPD models at forty steps, together with step-distilled variants of the base and joint OPD models at four steps, on GenEval \citep{ghosh2023geneval}, OCR accuracy~\citep{chen2023textdiffuser,cui2025paddleocr30technicalreport}, DeQA \citep{you2025deqa}, and PickScore \citep{kirstain2023pickscore}, combined into a single Average score. Tab.~\ref{tab:main_results} reports these four measures and Average for both SD3.5-Medium and FLUX.1-dev under the same protocol, and Tab.~\ref{tab:general_rewards} additionally reports CLIPScore~\citep{xu2024spectralclip}, HPSv2.1~\citep{wu2023human}, Aesthetic~\citep{AVA}, ImageReward~\citep{xu2023imagereward}, and UnifiedReward~\citep{unifiedreward} for the SD3.5-Medium rows, outside the Average score. See App.~\ref{sec:app:evaluation} for the Average formula, the generated image counts, and the scoring procedures.

\begin{figure}[t]
\centering
\includegraphics[width=\textwidth]{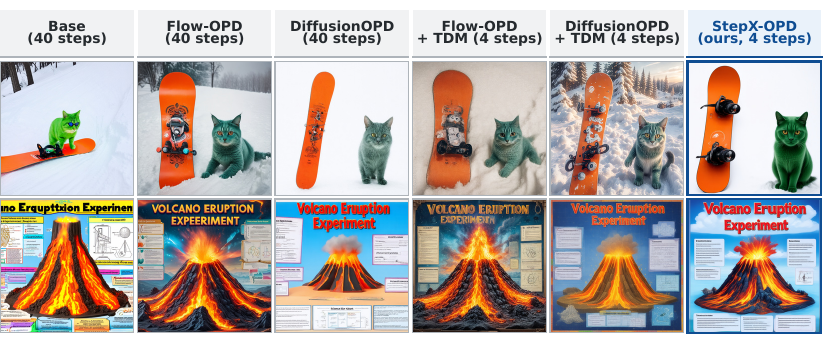}
\caption{Qualitative comparison result from benchmark.}
\label{fig:qualitative-comparison}
\end{figure}

\begin{figure}[t]
\centering
\includegraphics[width=1\linewidth]{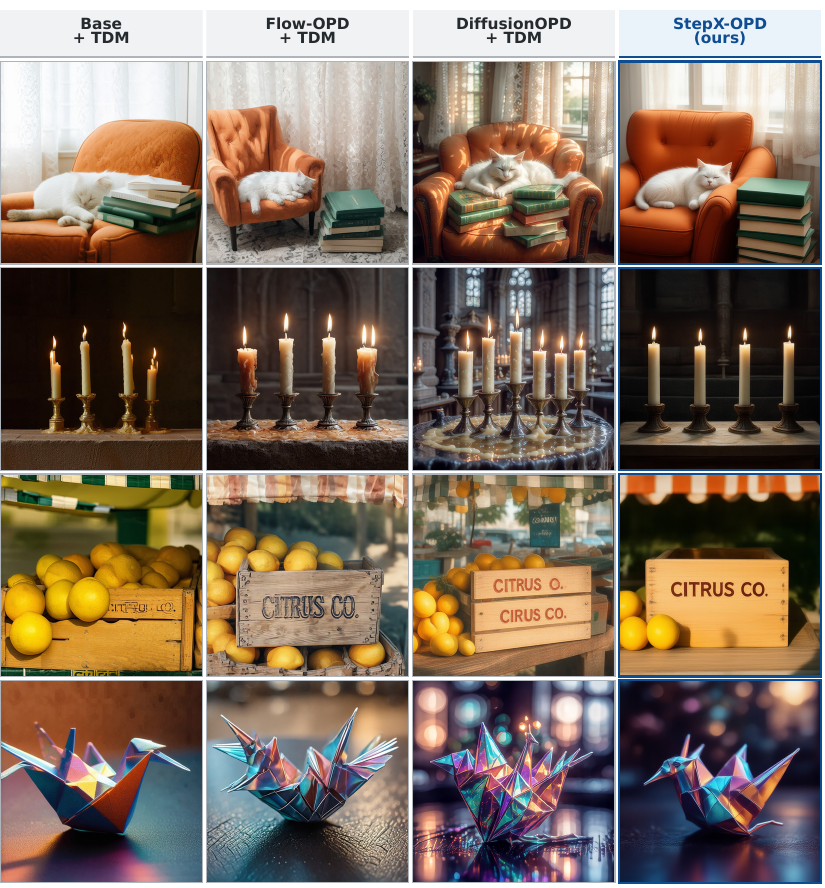}
\caption{Qualitative comparison result from generated prompt.}
\label{fig:quali-e}
\end{figure}

\noindent\textbf{Implementation details.} For the SD3.5-Medium four-step distillation main run, we adapt the transformer with rank-32 LoRA student and fake adapters, trained at learning rates $2\times10^{-6}$ and $1\times10^{-5}$ respectively, on four-step on-policy rollouts. Multi-teacher supervision covers the GenEval, OCR, and PickScore training sources, and \ours{} combines trajectory distribution matching with low-noise velocity regression against the routed teacher under static source routing, applied from step 500 onward, weighted by $w_{\text{v}}=0.5$, and gated on the batch-mean sampled time with a threshold of 0.5. For the SD3.5-Medium main run, we train with FSDP across 4 processes at a per-rank batch size of 64, giving a global batch size of 256, with 200-step optimizer warmups. The frozen initial student supplies the self-distillation target during the first 200 steps; afterward, source routing sends the whole mini-batch to one of three frozen DiffusionOPD teachers, with a static lookup routing the GenEval, OCR, and PickScore training sources to the GenEval, OCR, and aesthetic teachers, respectively.

\subsection{Main results}

\noindent\textbf{Quantitative.} To test whether joint training can deliver integrated capability at four steps, Tab.~\ref{tab:main_results} compares \ours{} with four-step TDM baselines and forty-step OPD models on both SD3.5-Medium and FLUX.1-dev. \ours{} reaches a four-step Average of 0.8901 on SD3.5-Medium and 0.8902 on FLUX.1-dev, against 0.8403 and 0.8761 for the next-best four-step baseline on each.

\noindent\textbf{Qualitative comparison.} Fig.~\ref{fig:qualitative-comparison} compares the forty-step and four-step models on one prompt from each of two benchmarks. On these two prompts, \ours{} reproduces the object relations the forty-step models render, while the other four-step columns shown here do not; \ours{} is also the only one of the six columns shown in this figure that spells the poster title correctly. Fig.~\ref{fig:quali-e} compares the four-step models on prompts written for this comparison. Further examples appear in App.~\ref{sec:app:qualitative}.

\subsection{Prompt-word recurrence}
\label{sec:exp:recurrence}

\begin{figure}[t]
\centering
\begin{minipage}[t]{0.48\textwidth}
\vspace{0pt}
\centering
\includegraphics[width=\linewidth]{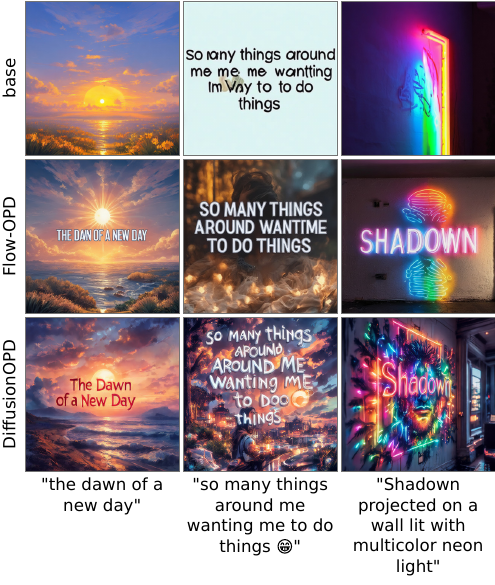}
\end{minipage}
\hfill
\begin{minipage}[t]{0.48\textwidth}
\vspace{0pt}
\centering
\includegraphics[width=\linewidth]{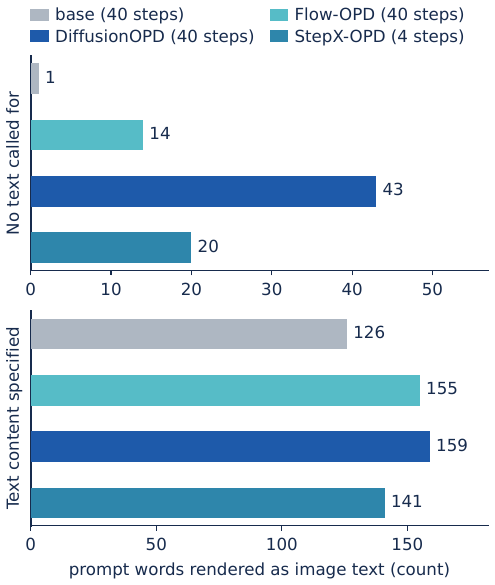}
\end{minipage}
\caption{Text that models render into the image. The left panel shows generations on the \emph{No text called for} prompt group; the right panel counts the generations whose detected on-image text reproduces a word from the prompt, on that same group and on \emph{Text content specified}, drawn on separate scales.}
\label{fig:spillover}
\end{figure}

To test whether capability integration is accompanied by unintended prompt-word rendering, Tab.~\ref{tab:recurrence_realbench} counts recurrence across eight models on prompts that do not request text, with seven models sampled at forty steps and \ours{} at four steps; Fig.~\ref{fig:spillover} repeats the comparison on a stricter prompt subset. The largest shift is on SD3.5-Medium, where recurrence rises from 21 images for the base model to 218 for DiffusionOPD. This shows that the recurrence measure exposes a text-rendering side effect associated with capability integration. \ours{}, trained with the DiffusionOPD teachers, reduces this count to 94.

\subsection{Ablation and analysis}
\label{sec:exp:ablation}

\noindent\textbf{Supervision-point construction.} To assess supervision-point construction, Tab.~\ref{tab:supervision_points} compares fixed-grid endpoints, randomized-grid endpoints, and randomized-grid interpolated re-noising. Replacing the fixed grid with the randomized one raises Average from 0.7090 to 0.7548, and moving the re-noising source from the segment endpoint to an interpolated point raises it further to 0.8240, with OCR rising from 0.4499 to 0.5776 to 0.7968 across the same three rows. Both changes therefore improve the aggregate score.

\begingroup
\let\experimenttable\table
\let\endexperimenttable\endtable
\renewenvironment{table}[1][]{\experimenttable[H]}{\endexperimenttable}

\noindent\textbf{Velocity-regression target source.} The additive initialization used in this section adds the SFT increment at full weight on top of the mean of the three specialist increments. Tab.~\ref{tab:vmse_grid} compares regressing toward the routed teacher's own velocity field against regressing toward a separate, dedicated quality teacher under low-noise coverage. The routed target reaches an Average of 0.8901, versus 0.8446 for the dedicated target.

\begin{table}[h]
\centering
\footnotesize
\setlength{\tabcolsep}{5pt}
\caption{Ablation of the three supervision-point constructions. Best results
are in bold.}
\label{tab:supervision_points}

\begin{tabular}{llrrrrr}
\toprule
Rollout grid & Re-noising source & GenEval $\uparrow$ & OCR $\uparrow$ & DeQA $\uparrow$ & PickScore $\uparrow$ & Average $\uparrow$ \\
\midrule
Fixed      & Segment endpoint        & 0.7230 & 0.4499 & 4.003 & 22.42 & 0.7090 \\
Randomized & Segment endpoint        & 0.7523 & 0.5776 & \textbf{4.129} & 22.46 & 0.7548 \\
\rowcolor{cyan!8}
Randomized & Interpolated & \textbf{0.8198} & \textbf{0.7968} & 4.076 & \textbf{22.47} & \textbf{0.8240} \\
\bottomrule
\end{tabular}
\end{table}
\noindent\textbf{Teacher selection.} To test how each mini-batch should pick which DiffusionOPD teacher supplies its distribution-matching target, Tab.~\ref{tab:source_routing} compares source routing, in which a mini-batch drawn from one training source is routed to that source's dedicated teacher, against two alternatives that break this correspondence: assigning a teacher at random per mini-batch, and averaging the three teachers' velocity fields into a single uniform target. Source routing reaches 0.8986 on OCR against 0.7388 for random assignment and 0.6738 for uniform averaging, the largest separation among the three rows, and source routing also reaches the highest GenEval, PickScore, and Average. Matching each mini-batch's teacher to its training source therefore matters most for the text-rendering objective.

\noindent\textbf{Initialization from released adapters.} To test dependence on student initialization, Tab.~\ref{tab:student_init} compares TDM and \ours{} from the released Flow-OPD and DiffusionOPD adapters. In each pair \ours{} uses the specialist teachers of the same family as the adapter, the Flow-OPD teachers from the Flow-OPD adapter and the DiffusionOPD teachers from the DiffusionOPD adapter. From the Flow-OPD adapter, \ours{} raises Average from 0.8345 to 0.8746, the larger gain of the two pairs, and \ours{} also reaches the higher score from the DiffusionOPD adapter.

\begin{table}[t]
\centering
\footnotesize
\caption{Ablation of how each mini-batch selects its teacher, reported as full-benchmark scores. Best results are in bold.}
\label{tab:source_routing}
\begin{tabular}{lrrrrr}
\toprule
Teacher selection & GenEval $\uparrow$ & OCR $\uparrow$ & DeQA $\uparrow$ & PickScore $\uparrow$ & Average $\uparrow$ \\
\midrule
\rowcolor{cyan!8}
Source routing        & \textbf{0.9442} & \textbf{0.8986} & 4.170          & \textbf{22.98} & \textbf{0.8901} \\
Random teacher         & 0.9241          & 0.7388          & \textbf{4.203} & 22.37          & 0.8410 \\
Uniform teacher average & 0.9292          & 0.6738          & 4.179          & 22.18          & 0.8230 \\
\bottomrule
\end{tabular}
\vspace{0pt}
\end{table}

\begin{table}[t]
\centering
\fontsize{7}{8.5}\selectfont
\setlength{\tabcolsep}{1pt}
\begin{minipage}[t]{0.485\textwidth}
\vspace{0pt}
\centering
\caption{Comparison of four configurations, grouped by velocity-regression noise coverage and by the model that supplies the regression target. Best results are in bold.}
\label{tab:vmse_grid}
\begin{tabular}{llrrrrr}
\toprule
Coverage & Target & Gen $\uparrow$ & OCR $\uparrow$ & DeQA $\uparrow$ & Pick $\uparrow$ & Avg $\uparrow$ \\
\midrule
\rowcolor{cyan!8}
\cellcolor{white} \multirow{2}{*}{Low-noise}  & Routed     & \textbf{0.9442} & \textbf{0.8986} & \textbf{4.170} & \textbf{22.98} & \textbf{0.8901} \\
                            & Dedicated  & 0.8822 & 0.8127 & 4.057 & 22.67 & 0.8446 \\
\addlinespace[2pt]
\multirow{2}{*}{Full-range} & Routed     & 0.8894    & 0.8341    & 4.121   & 22.73   & 0.8555    \\
                            & Dedicated  & 0.8535 & 0.7832 & 4.115 & 22.03 & 0.8268 \\
\bottomrule
\end{tabular}

\end{minipage}
\hfill
\begin{minipage}[t]{0.485\textwidth}
\vspace{0pt}
\centering
\caption{Comparison of distillation methods across two student initializations. Best results
are in bold.}
\label{tab:student_init}
\begin{tabular}{llrrrrr}
\toprule
Init. & Method & Gen $\uparrow$ & OCR $\uparrow$ & DeQA $\uparrow$ & Pick $\uparrow$ & Avg $\uparrow$ \\
\midrule
 & TDM & 0.875 & 0.751 & \textbf{4.217} & 22.60 & 0.8345 \\
\rowcolor{cyan!8}
\cellcolor{white}\multirow{-2}{*}{Flow-OPD} & \ours{} & \textbf{0.909} & \textbf{0.897} & 4.103 & \textbf{22.68} & \textbf{0.8746} \\
\addlinespace[2pt]
 & TDM & 0.838 & 0.833 & 3.952 & \textbf{23.41} & 0.8403 \\
\rowcolor{cyan!8}
\cellcolor{white}\multirow{-2}{*}{DiffusionOPD} & \ours{} & \textbf{0.906} & \textbf{0.893} & \textbf{4.056} & 22.92 & \textbf{0.8727} \\
\bottomrule
\end{tabular}

\end{minipage}
\end{table}

\noindent\textbf{Velocity-regression term.} To examine how the velocity-regression term should be applied, Tab.~\ref{tab:vmse_coverage_start_step} compares omitting it against applying it from step 0 and from step 500. All three rows share the same SD3.5-Medium four-step student, SFT-additive initialization, three DiffusionOPD specialists, fixed source routing, source mixture, randomized-grid interpolated re-noising, optimizer configuration, training-step count, and evaluation protocol, with FSDP across 4 processes at a per-rank batch of 64 (global batch 256); this controls the update budget and recipe, so the routed multi-teacher row without velocity regression is the matched baseline. Omitting the term reaches an Average of 0.8722, and applying it from step 0 and from step 500 reaches 0.8895 and 0.8901.

\noindent\textbf{SFT-increment merge weight.} To compare how the SFT increment enters the student initialization, Tab.~\ref{tab:sft_merge} tests the SFT increment on its own, the base merge, a four-model mean that averages the SFT and the three specialist increments together, and the additive merge defined above. See App.~\ref{sec:app:evaluation} for the expressions. Each row evaluates the initialization weights themselves, before \ours{} training is applied. The additive variant reaches the highest Average, 0.8348 versus 0.8219 for the next-best row. This supports the additive merge as the strongest initialization variant tested.

\begin{table}[t]
\centering
\fontsize{7}{8.5}\selectfont
\setlength{\tabcolsep}{1pt}
\begin{minipage}[t]{0.485\textwidth}
\vspace{0pt}
\centering
\caption{Ablation of the velocity-regression term, covering whether it is applied, and the step from which it is applied. Best results are in bold.}
\label{tab:vmse_coverage_start_step}
\begin{tabular}{lrrrrr}
\toprule
Velocity regression & Gen $\uparrow$ & OCR $\uparrow$ & DeQA $\uparrow$ & Pick $\uparrow$ & Avg $\uparrow$ \\
\midrule
w/o regression    & 0.9434 & 0.8656 & \textbf{4.178} & 21.95 & 0.8722 \\
from step 0       & \textbf{0.9444} & 0.8950 & 4.177 & 22.96 & 0.8895 \\
\rowcolor{cyan!8}
from step 500     & 0.9442 & \textbf{0.8986} & 4.170 & \textbf{22.98} & \textbf{0.8901} \\
\bottomrule
\end{tabular}

\end{minipage}
\hfill
\begin{minipage}[t]{0.485\textwidth}
\vspace{0pt}
\centering
\caption{Ablation of SFT-merge variants of the student initialization, reported as full-benchmark scores. Best results
are in bold.}
\label{tab:sft_merge}
\begin{tabular}{lrrrrr}
\toprule
Initialization & Gen $\uparrow$ & OCR $\uparrow$ & DeQA $\uparrow$ & Pick $\uparrow$ & Avg $\uparrow$ \\
\midrule
SFT                  & 0.9007 & 0.6343 & 4.024 & 22.63 & 0.8025 \\
Merge                & 0.8664 & \textbf{0.7296} & 4.101 & 22.65 & 0.8219 \\
SFT Merge~(mean)     & 0.8758 & 0.7054 & \textbf{4.116} & 22.65 & 0.8189 \\
\rowcolor{cyan!8}
SFT Merge~(additive) & \textbf{0.9273} & 0.7294 & 4.050 & \textbf{22.68} & \textbf{0.8348} \\
\bottomrule
\end{tabular}

\end{minipage}
\end{table}

\endgroup

\section{Conclusion}
Multi-capability, few-step text-to-image models are typically built in two stages: a full-step model consolidates capabilities, then serves as a fixed teacher for compression. This lengthens training, and the capabilities gathered in the first stage do not fully survive into the second. \ours{} merges the two stages, learning capabilities directly on the few-step trajectory the deployed model runs and removing the intermediate full-step teacher. A reverse-KL decomposition motivates velocity regression on the trajectory's low-noise part. Capability integration also leads models to render prompt words into images on prompts that neither request nor forbid text, and we give a variant of \ours{} that reduces this behavior. Evaluated on SD3.5-Medium and FLUX.1-dev, \ours{} reaches the highest Average among the four-step methods on both and exceeds the forty-step Flow-OPD, and on SD3.5-Medium it trains with about 37\% fewer GPU-hours than sequential DiffusionOPD followed by step distillation.

\bibliography{iclr2027_conference}
\bibliographystyle{iclr2027_conference}

\clearpage
\begin{center}
{\LARGE\sc Appendix\par}
\end{center}
\vskip 0.3in
\appendix

\startcontents[appendix]
\section*{Contents of Appendix}
\setcounter{tocdepth}{2}
\printcontents[appendix]{}{1}{}
\vspace{1em}
\newpage

\section{Theoretical Analysis}
\label{sec:app:theory}

\subsection{Setting and Assumptions} \label{sec:theory:setting}
DMD matches distributions after forward noising, and TDM applies this objective to states along the student's trajectory \citep{yin2024dmd,luo2025tdm}. OPD matches the teacher's behavior at states generated by the student \citep{li2026diffusionopd}. We fix a text condition and relate these objectives on the same noised states.
\subsubsection{Sampling and Forward Laws} \label{sec:theory:sampling}
Given a random rollout grid, first select one of its $K$ segments uniformly at random and then draw the base time uniformly within that segment:
\begin{equation}
\mathcal T=(1=t_K>\dots>t_0=0),\qquad
i\sim\Unif\{1,\dots,K\},\qquad b\mid i\sim\Unif[t_{i-1},t_i). \label{eq:theory:grid_base}
\end{equation}
Interpolate the two endpoint rollout states to obtain the state at the sampled base:
\begin{equation}
X_b=\frac{(t_i-b)X_{t_{i-1}}+(b-t_{i-1})X_{t_i}}{t_i-t_{i-1}}. \label{eq:theory:base}
\end{equation}
The analysis below conditions on the sampled $(\mathcal T,b)$, with $b<t_i\le1$. Segment lengths vary under a random grid, so this two-stage draw does not make $b$ itself uniform on $(0,1)$; conditioning on $(\mathcal T,b)$ means this marginal law does not affect the conclusions that follow. When $t_i=1$, the uniform draw satisfies $T<1$ almost surely, so every velocity and identity is evaluated below $1$. Values assigned at the sampling endpoints do not affect the Lebesgue expectations. Let $q_b$ be the conditional distribution of $X_b$, and let $p_b$ be the teacher distribution at the same time. All laws are on $\mathbb R^d$, and $\KL(q\|p)$ denotes relative entropy. For $b<t<1$, apply the same forward noising kernel to both laws:
\begin{equation}
\begin{aligned}
k_{t|b}(\cdot\mid x)&=\mathcal N(a_{t|b}x,s_{t|b}^2 I),\\
a_{t|b}&=\frac{1-t}{1-b},\qquad
s_{t|b}^2=t^2-a_{t|b}^2b^2.
\end{aligned} \label{eq:theory:kernel}
\end{equation}
\begin{equation}
q_t=k_{t|b}q_b,\qquad p_t=k_{t|b}p_b. \label{eq:theory:forward_laws}
\end{equation}
Here $kr=\int k(\cdot\mid x)r(\dd x)$, and $q_t$ is the forward-noised law from $q_b$. Next sample the supervision time and apply the exact forward SDE transition from $b$ to $T$:
\begin{equation}
\begin{aligned}
&T\mid\mathcal T,b\sim\Unif[b,t_i],\qquad
\epsilon\sim\mathcal N(0,I),\\
&Z_T=a_{T|b}X_b+s_{T|b}\epsilon,\qquad
Z_T\mid T,\mathcal T,b\sim q_T.
\end{aligned} \label{eq:theory:renoise}
\end{equation}
Unless stated otherwise, expectations are conditional on $(\mathcal T,b)$. The weight used throughout is
\begin{equation}
w_b(t)=(t_i-t)\frac{1-t}{t},\qquad b<t<t_i. \label{eq:theory:weight}
\end{equation}
\subsubsection{Assumptions} \label{sec:theory:assumptions}
The statements below use the following assumptions at the fixed grid and base.
\begin{enumerate}[label=\textbf{A\arabic*.},ref=A\arabic*,leftmargin=*]
\item\label{as:theory:sampling} \textbf{Conditional sampling and common kernel.}
Given $(\mathcal T,b)$, $T$ is independent of $X_b$, and $\epsilon$ is independent of $(X_b,T)$. Both distributions evolve through the kernel in~\eqref{eq:theory:kernel}, and the samples follow~\eqref{eq:theory:renoise}.
\item\label{as:theory:regularity} \textbf{Regularity and the base limit.}
The laws $q_b,p_b$ have finite second moments and $\KL(q_b\|p_b)<\infty$. Their Gaussian-smoothed densities satisfy the smoothness and tail conditions of the relative de Bruijn identity below. In particular, differentiation under the integral and integration by parts are valid, with vanishing boundary terms. The function $K(t)=\KL(q_t\|p_t)$ is continuously differentiable on $(b,t_i)$ and satisfies $K(t)\to\KL(q_b\|p_b)$ as $t\downarrow b$.
\item\label{as:theory:teacher} \textbf{Teacher consistency.}
Let $v_\phi$ denote the teacher network velocity. Assume $v_\phi(z,t)=v_p(z,t)$ for $b<t<t_i$, where $v_p$ is the distribution-induced velocity in~\eqref{eq:theory:velocity}. The student network velocity is the measurable function $v_\theta$. The teacher network velocity $v_\phi$ is the velocity field produced by the teacher together with its guidance configuration, and $p$ is defined to be the distribution induced by this guided field, following the convention used throughout distillation methods built on classifier-free guidance.
\item\label{as:theory:integrability} \textbf{Weighted integrability.}
The weighted squared discrepancies defining $O_b$ and $R_b$ in~\eqref{eq:theory:O} and~\eqref{eq:theory:R} are integrable.
\end{enumerate}
\subsection{Preliminary Identities} \label{sec:theory:preliminaries}
\subsubsection{Semigroup of the Forward Kernel} \label{sec:theory:semigroup}
For $b<s<t<1$, the Gaussian kernels satisfy
\begin{equation}
k_{t|s}k_{s|b}=k_{t|b}. \label{eq:theory:semigroup}
\end{equation}
Indeed, composition gives a Gaussian kernel, whose mean coefficient and variance are
\begin{align}
a_{t|s}a_{s|b}
&=\frac{1-t}{1-s}\frac{1-s}{1-b}=a_{t|b},\label{eq:theory:semigroup_mean}\\
a_{t|s}^2s_{s|b}^2+s_{t|s}^2
&=a_{t|s}^2\big(s^2-a_{s|b}^2b^2\big)
  +t^2-a_{t|s}^2s^2\notag\\
&=t^2-a_{t|b}^2b^2=s_{t|b}^2.\label{eq:theory:semigroup_variance}
\end{align}
Consequently, $q_t=k_{t|s}q_s$ and $p_t=k_{t|s}p_s$. Both smoothed laws therefore follow the same forward evolution.
\subsubsection{Euler Interpolation and Exact Re-noising} \label{sec:theory:interpolation}
\begin{lemma}[Interpolation is an Euler sub-step]\label{lem:interp}
Suppose the rollout endpoints satisfy the Euler update of Eq.~\ref{eq:euler_step} that produced them, $z_{\mathrm{end}}=z_{\mathrm{start}}+(\sigma_\ell-\sigma_u)\,v_\theta(z_{\mathrm{start}},\sigma_u)$. Then the interpolated state of Eq.~\ref{eq:xbase} satisfies
\begin{equation*}
x_{\mathrm{base}}=z_{\mathrm{start}}+(\tau_{\mathrm{base}}-\sigma_u)\,v_\theta(z_{\mathrm{start}},\sigma_u),
\end{equation*}
so $x_{\mathrm{base}}$ is exactly the state reached by inserting $\tau_{\mathrm{base}}$ into the grid and taking one Euler step from $\sigma_u$. Moreover, under the sampling independence assumed in Assumption~\ref{as:theory:sampling}, Gaussian re-noising from $x_{\mathrm{base}}$ to any later time carries $x_{\mathrm{base}}$'s own conditional distribution forward exactly, whatever that distribution is, with no approximation.
\end{lemma}
\begin{proof}[Proof of Lemma~\ref{lem:interp}]
Fix the sampled grid and base. By hypothesis, the two rollout endpoints satisfy the Euler update
\begin{equation}
X_{t_{i-1}}=X_{t_i}+(t_{i-1}-t_i)v_\theta(X_{t_i},t_i).
\label{eq:theory:euler_step}
\end{equation}
Substitute~\eqref{eq:theory:euler_step} into~\eqref{eq:theory:base} and collect the coefficients of $X_{t_i}$:
\begin{align}
X_b
&=\frac{(t_i-b)\big[X_{t_i}+(t_{i-1}-t_i)v_\theta(X_{t_i},t_i)\big]
 +(b-t_{i-1})X_{t_i}}{t_i-t_{i-1}}\notag\\
&=X_{t_i}+\frac{(t_i-b)(t_{i-1}-t_i)}{t_i-t_{i-1}}
 v_\theta(X_{t_i},t_i)\notag\\
&=X_{t_i}+(b-t_i)v_\theta(X_{t_i},t_i).
\label{eq:theory:interpolation_proof}
\end{align}
The sum of those coefficients is $[(t_i-b)+(b-t_{i-1})]/(t_i-t_{i-1})=1$, and $(t_{i-1}-t_i)/(t_i-t_{i-1})=-1$ gives the last line, so~\eqref{eq:theory:interpolation_proof} is exactly the state reached by inserting $b$ into the grid and advancing from $t_i$ to $b$ by one Euler step.
Inserting $b$ below $t_i$ leaves the already computed states at $t_K,\ldots,t_i$ unchanged.
When $b=t_{i-1}$, \eqref{eq:theory:euler_step} gives the lower endpoint directly, matching~\eqref{eq:theory:interpolation_proof} at that boundary.
Under the conditional independence assumed in Assumption~\ref{as:theory:sampling}, re-noising from this base has exactly the law $k_{t|b}q_b$ for $b<t<1$, where $q_b$ is the actual conditional law of the interpolated state. For re-noising, condition on $X_b=x$. Independence of $\epsilon$ makes $Z_t=a_{t|b}x+s_{t|b}\epsilon$ Gaussian with kernel $k_{t|b}(\cdot\mid x)$.
Thus, for every Borel set $A$,
\begin{equation}
\begin{aligned}
\mathbb P(Z_t\in A\mid\mathcal T,b)
&=\int\mathbb P(Z_t\in A\mid X_b=x,\mathcal T,b)\,q_b(\dd x)\\
&=\int k_{t|b}(A\mid x)\,q_b(\dd x)
=(k_{t|b}q_b)(A).
\end{aligned}
\label{eq:theory:renoise_law_proof}
\end{equation}
For $b<s<t<1$, the Chapman--Kolmogorov identity~\eqref{eq:theory:semigroup} also gives
\begin{equation}
k_{t|s}(k_{s|b}q_b)=(k_{t|s}k_{s|b})q_b=k_{t|b}q_b.
\label{eq:theory:renoise_consistency}
\end{equation}
so adding noise in two stages and adding it directly give the same marginal law.
\end{proof}
\paragraph{Relation to the continuous ODE.}
The interpolation identity is exact for the piecewise-linear Euler trajectory. For a continuous solution of $\dot x(t)=v_\theta(x(t),t)$, the Euler sub-step is a numerical approximation. Even when both endpoints lie on the same exact ODE solution, linear interpolation reproduces every intermediate state exactly only when that solution segment is affine, equivalently when its velocity is constant along the segment. Indeed, agreement with the endpoint chord at every intermediate time makes the trajectory affine, and the converse follows by interpolation. The Gaussian transition above is exact for the actual base law $q_b$, including any approximation error in that law relative to a continuous ODE marginal.
\paragraph{Coefficients of the re-noising step used in the main text.}
Sec.~\ref{sec:method} moves from $\tau_{\text{base}}$ to $\tau$ along the forward path $z_t=\alpha_t x_0+\sigma_t\epsilon$ with $\alpha_t=1-t$ and $\sigma_t=t$, writing this transition as $z_\tau=\kappa\,x_{\text{base}}+\beta\,\epsilon$ in Eq.~\ref{eq:renoise}. This step is the kernel of Eq.~\ref{eq:theory:kernel} evaluated at $t=\tau$ and $b=\tau_{\text{base}}$, so $\kappa=a_{\tau|\tau_{\text{base}}}$ and $\beta=s_{\tau|\tau_{\text{base}}}$, and the derivation below writes the two coefficients out directly in the $\alpha_t,\sigma_t$ notation that Sec.~\ref{sec:method} uses. Requiring $z_\tau=\kappa\,x_{\text{base}}+\beta\,\epsilon'$, with $\epsilon'$ independent of the noise already carried by $x_{\text{base}}$, to share the same law as $\alpha_\tau x_0+\sigma_\tau\epsilon$ fixes both coefficients by matching moments. Matching the coefficient of $x_0$ gives $\kappa\,\alpha_{\tau_{\text{base}}}=\alpha_\tau$, hence
\begin{equation}
\kappa=\frac{\alpha_\tau}{\alpha_{\tau_{\text{base}}}},
\label{eq:app:renoise_kappa}
\end{equation}
and matching the variance contributed by the Gaussian noise gives $(\kappa\,\sigma_{\tau_{\text{base}}})^2+\beta^2=\sigma_\tau^2$, hence
\begin{equation}
\beta^2=\sigma_\tau^2-(\kappa\,\sigma_{\tau_{\text{base}}})^2.
\label{eq:app:renoise_beta}
\end{equation}
Under the ordering $\tau\ge\tau_{\text{base}}$ that Eq.~\ref{eq:renoise} assumes, this right-hand side is nonnegative, so the $\max(\cdot,0)$ that Eq.~\ref{eq:renoise} applies before the square root only guards against floating-point round-off at that boundary, not against a genuine sign change.
\subsubsection{Velocity and Score} \label{sec:theory:velocity}
For $r\in\{q,p\}$, define the velocity induced by the forward law:
\begin{equation}
v_r(z,t)=-\frac{z}{1-t}-\frac{t}{1-t}\nabla_z\log r_t(z). \label{eq:theory:velocity}
\end{equation}
Subtracting the two definitions cancels the common drift. Thus,
\begin{align}
v_q(z,t)-v_p(z,t)
&=-\frac{t}{1-t}\big(\nabla_z\log q_t(z)-\nabla_z\log p_t(z)\big), \label{eq:theory:velocity_score}\\
\|\nabla_z\log q_t(z)-\nabla_z\log p_t(z)\|^2
&=\frac{(1-t)^2}{t^2}\|v_q(z,t)-v_p(z,t)\|^2. \label{eq:theory:score_square}
\end{align}
\subsubsection{Relative de Bruijn Identity} \label{sec:theory:debruijn}
Let $u>0$ be a Gaussian variance parameter and define
\begin{equation}
\widehat q_u=q_b*\mathcal N(0,uI),\qquad
\widehat p_u=p_b*\mathcal N(0,uI). \label{eq:theory:heat_laws}
\end{equation}
Theorem~1 of \citet{lyu2009score} gives the relative KL dissipation identity under Gaussian smoothing, also called a generalized de Bruijn identity:
\begin{equation}
\frac{\dd}{\dd u}\KL(\widehat q_u\|\widehat p_u)
=-\frac12\E_{Y\sim\widehat q_u}
\|\nabla_y\log\widehat q_u(Y)-\nabla_y\log\widehat p_u(Y)\|^2. \label{eq:theory:lyu}
\end{equation}
We use this established identity directly. The change of variables for our forward kernel is given in Proposition~\ref{prop:theory:dissipation}.
\subsubsection{A Uniform Random Integration Endpoint} \label{sec:theory:endpoint}
For any measurable $h\ge0$ and $T\sim\Unif[b,t_i]$, Tonelli's theorem gives
\begin{align}
\E_T\!\left[\int_b^T h(t)\,\dd t\right]
&=\frac1{t_i-b}\int_b^{t_i}\int_b^\tau h(t)\,\dd t\,\dd\tau\notag\\
&=\frac1{t_i-b}\int_b^{t_i}\int_t^{t_i}h(t)\,\dd\tau\,\dd t\notag\\
&=\frac1{t_i-b}\int_b^{t_i}(t_i-t)h(t)\,\dd t\notag\\
&=\E_T\big[(t_i-T)h(T)\big]. \label{eq:theory:endpoint_identity}
\end{align}
The factor $1/(t_i-b)$ is already included in the uniform expectation in the last line.
\subsection{Main Results and Proofs} \label{sec:theory:results}
For an Euler rollout, Lemma~\ref{lem:interp} identifies $q_b$ as the law of an interpolated Euler state and justifies its exact Gaussian smoothing. The KL results below apply to this actual base law under the stated assumptions.
Define
\begin{equation}
K(t)=\KL(q_t\|p_t),\qquad
m(t)=\E_{Z\sim q_t}\|v_q(Z,t)-v_p(Z,t)\|^2. \label{eq:theory:K_m}
\end{equation}
The three quantities used in the main theorem are
\begin{align}
D_b&=\E_T\,\KL(q_T\|p_T),\label{eq:theory:D}\\
O_b&=\E\big[w_b(T)\|v_\theta(Z_T,T)-v_\phi(Z_T,T)\|^2\big],\label{eq:theory:O}\\
R_b&=\E\big[w_b(T)\|v_\theta(Z_T,T)-v_q(Z_T,T)\|^2\big].\label{eq:theory:R}
\end{align}
\begin{proposition}[Dissipation under the forward kernel] \label{prop:theory:dissipation}
Under Assumptions~\ref{as:theory:sampling} and~\ref{as:theory:regularity}, for $b<t<t_i$,
\begin{equation}
K'(t)=-\frac{1-t}{t}m(t). \label{eq:theory:dissipation}
\end{equation}
\end{proposition}
\begin{proof}
For this proof, write $a(t)=a_{t|b}$. The rescaled state and its Gaussian noise variance are
\begin{equation}
\begin{aligned}
Y_t&=\frac{Z_t}{a(t)}=X_b+\sqrt{\lambda_b(t)}\,\epsilon,\\
\lambda_b(t)&=\frac{s_{t|b}^2}{a(t)^2}
=\frac{(1-b)^2t^2}{(1-t)^2}-b^2.
\end{aligned} \label{eq:theory:variance_time}
\end{equation}
The kernel in~\eqref{eq:theory:kernel} is equivalent to Gaussian smoothing with variance $\lambda_b(t)$ followed by multiplication by $a(t)$. The resulting densities are
\begin{equation}
q_t(z)=a(t)^{-d}\widehat q_{\lambda_b(t)}(z/a(t)),\qquad
p_t(z)=a(t)^{-d}\widehat p_{\lambda_b(t)}(z/a(t)). \label{eq:theory:scaled_densities}
\end{equation}
Using the substitution $y=z/a(t)$ and $\dd z=a(t)^d\dd y$, the common scaling cancels in KL:
\begin{align}
K(t)
&=\int a(t)^{-d}\widehat q_{\lambda_b(t)}(z/a(t))
\log\frac{\widehat q_{\lambda_b(t)}(z/a(t))}{\widehat p_{\lambda_b(t)}(z/a(t))}\,\dd z\notag\\
&=\int\widehat q_{\lambda_b(t)}(y)
\log\frac{\widehat q_{\lambda_b(t)}(y)}{\widehat p_{\lambda_b(t)}(y)}\,\dd y\notag\\
&=\KL(\widehat q_{\lambda_b(t)}\|\widehat p_{\lambda_b(t)}). \label{eq:theory:KL_scaling}
\end{align}
The spatial chain rule applied to~\eqref{eq:theory:scaled_densities} gives, for $r\in\{q,p\}$,
\begin{equation}
\nabla_z\log r_t(z)=a(t)^{-1}
\nabla_y\log\widehat r_{\lambda_b(t)}(z/a(t)). \label{eq:theory:score_scaling}
\end{equation}
Since $Z=a(t)Y$ has law $q_t$ when $Y\sim\widehat q_{\lambda_b(t)}$, this implies
\begin{equation}
\begin{aligned}
&\E_{Y\sim\widehat q_{\lambda_b(t)}}
\|\nabla\log\widehat q_{\lambda_b(t)}(Y)-\nabla\log\widehat p_{\lambda_b(t)}(Y)\|^2\\
&\qquad=a(t)^2\E_{Z\sim q_t}
\|\nabla\log q_t(Z)-\nabla\log p_t(Z)\|^2.
\end{aligned} \label{eq:theory:fisher_scaling}
\end{equation}
Differentiating the variance parameter in~\eqref{eq:theory:variance_time} gives
\begin{align}
\lambda'_b(t)
&=(1-b)^2\left(\frac{2t}{(1-t)^2}+\frac{2t^2}{(1-t)^3}\right)\notag\\
&=\frac{2(1-b)^2t}{(1-t)^3}. \label{eq:theory:variance_derivative}
\end{align}
Apply the time chain rule to~\eqref{eq:theory:KL_scaling}, substitute Lyu's identity~\eqref{eq:theory:lyu}, and then use~\eqref{eq:theory:fisher_scaling}:
\begin{align}
K'(t)
&=\lambda'_b(t)\left.\frac{\dd}{\dd u}\KL(\widehat q_u\|\widehat p_u)\right|_{u=\lambda_b(t)}\notag\\
&=-\frac{\lambda'_b(t)}2\E_{Y\sim\widehat q_{\lambda_b(t)}}
\|\nabla\log\widehat q_{\lambda_b(t)}(Y)-\nabla\log\widehat p_{\lambda_b(t)}(Y)\|^2\notag\\
&=-\frac{\lambda'_b(t)a(t)^2}2\E_{Z\sim q_t}
\|\nabla\log q_t(Z)-\nabla\log p_t(Z)\|^2. \label{eq:theory:dissipation_chain}
\end{align}
The coefficient reduces to
\begin{equation}
\frac{\lambda'_b(t)a(t)^2}2
=\frac{(1-b)^2t}{(1-t)^3}\frac{(1-t)^2}{(1-b)^2}
=\frac{t}{1-t}. \label{eq:theory:dissipation_coefficient}
\end{equation}
Finally substitute the velocity--score relation~\eqref{eq:theory:score_square}:
\begin{align}
K'(t)
&=-\frac{t}{1-t}\E_{Z\sim q_t}
\|\nabla\log q_t(Z)-\nabla\log p_t(Z)\|^2\notag\\
&=-\frac{t}{1-t}\frac{(1-t)^2}{t^2}m(t)\notag\\
&=-\frac{1-t}{t}m(t). \label{eq:theory:dissipation_velocity}
\end{align}
\end{proof}
\begin{theorem}[Decomposition, bound, and self-consistency] \label{thm:theory:decomposition}
Under Assumptions~\ref{as:theory:sampling} through~\ref{as:theory:integrability}, at the fixed sampled $(\mathcal T,b)$ with $t_i\le1$, the exact decomposition is
\begin{equation}
\boxed{
\KL(q_b\|p_b)
=D_b+\E\big[w_b(T)\|v_q(Z_T,T)-v_p(Z_T,T)\|^2\big]
}. \label{eq:theory:decomposition}
\end{equation}
For every $\eta>0$, it yields
\begin{equation}
\boxed{\KL(q_b\|p_b)\le D_b+(1+\eta)O_b+(1+\eta^{-1})R_b}. \label{eq:theory:bound_eta}
\end{equation}
In particular, setting $\eta=1$ gives
\begin{equation}
\boxed{\KL(q_b\|p_b)\le D_b+2O_b+2R_b}. \label{eq:theory:bound}
\end{equation}
If $R_b=0$, the decomposition becomes
\begin{equation}
\boxed{\KL(q_b\|p_b)=D_b+O_b}. \label{eq:theory:consistent}
\end{equation}
\end{theorem}
\begin{proof}
Fix $b<s<\tau<t_i$. Integrating Proposition~\ref{prop:theory:dissipation} on $[s,\tau]$ gives
\begin{equation}
K(s)-K(\tau)=\int_s^\tau\frac{1-t}{t}m(t)\,\dd t. \label{eq:theory:integrated_interior}
\end{equation}
The integrand is nonnegative. Let $s\downarrow b$. Assumption~\ref{as:theory:regularity} gives the limit of $K(s)$, and monotone convergence gives the limit of the integral:
\begin{equation}
\KL(q_b\|p_b)=K(\tau)+\int_b^\tau\frac{1-t}{t}m(t)\,\dd t. \label{eq:theory:integrated_base}
\end{equation}
Both terms on the right are nonnegative, and their sum is finite. Now average over $\tau=T\sim\Unif[b,t_i]$ and apply~\eqref{eq:theory:endpoint_identity} with $h(t)=(1-t)m(t)/t$:
\begin{align}
\KL(q_b\|p_b)
&=\E_T K(T)+\E_T\!\left[\int_b^T\frac{1-t}{t}m(t)\,\dd t\right]\notag\\
&=D_b+\E_T\!\left[(t_i-T)\frac{1-T}{T}m(T)\right]\notag\\
&=D_b+\E_T\!\left[w_b(T)\E_{Z_T\mid T}
\|v_q(Z_T,T)-v_p(Z_T,T)\|^2\right]\notag\\
&=D_b+\E\big[w_b(T)\|v_q(Z_T,T)-v_p(Z_T,T)\|^2\big]. \label{eq:theory:averaged_decomposition}
\end{align}
The first equality averages~\eqref{eq:theory:integrated_base}. The second uses the uniform-endpoint identity and the definition of $D_b$. The third substitutes the definitions of $w_b$ and $m$, with the conditional law from~\eqref{eq:theory:renoise}. The last equality is the law of total expectation. This proves~\eqref{eq:theory:decomposition}.

For the bound, omit the shared arguments $(Z_T,T)$. By Assumption~\ref{as:theory:teacher}, $v_p=v_\phi$. Write $A=v_\theta-v_\phi$ and $B=v_q-v_\theta$. For $\eta>0$, use $2\langle A,B\rangle\le\eta\|A\|^2+\eta^{-1}\|B\|^2$ to obtain
\begin{align}
\|v_q-v_p\|^2
&=\|v_q-v_\phi\|^2=\|A+B\|^2\notag\\
&=\|A\|^2+2\langle A,B\rangle+\|B\|^2\notag\\
&\le(1+\eta)\|A\|^2+(1+\eta^{-1})\|B\|^2\notag\\
&=(1+\eta)\|v_\theta-v_\phi\|^2
 +(1+\eta^{-1})\|v_\theta-v_q\|^2. \label{eq:theory:triangle}
\end{align}
Multiply by $w_b(T)\ge0$, take expectations, and substitute~\eqref{eq:theory:decomposition}:
\begin{align}
\KL(q_b\|p_b)-D_b
&=\E\big[w_b(T)\|v_q-v_p\|^2\big]\notag\\
&\le(1+\eta)\E\big[w_b(T)\|v_\theta-v_\phi\|^2\big]\notag\\
&\quad +(1+\eta^{-1})\E\big[w_b(T)\|v_\theta-v_q\|^2\big]\notag\\
&=(1+\eta)O_b+(1+\eta^{-1})R_b. \label{eq:theory:bound_proof}
\end{align}
This proves~\eqref{eq:theory:bound_eta}; setting $\eta=1$ gives~\eqref{eq:theory:bound}. If $R_b=0$, its nonnegative integrand is zero almost surely:
\begin{equation}
w_b(T)\|v_\theta-v_q\|^2=0\qquad\text{a.s.} \label{eq:theory:zero_residual}
\end{equation}
Where $w_b(T)>0$, this implies $v_\theta=v_q$. Where the weight is zero, the weighted discrepancies below vanish. Using teacher consistency again gives
\begin{equation}
\begin{aligned}
\E\big[w_b(T)\|v_q-v_p\|^2\big]
&=\E\big[w_b(T)\|v_\theta-v_p\|^2\big]\\
&=\E\big[w_b(T)\|v_\theta-v_\phi\|^2\big]=O_b.
\end{aligned} \label{eq:theory:consistent_proof}
\end{equation}
Substituting this equality into~\eqref{eq:theory:decomposition} proves~\eqref{eq:theory:consistent}.
\end{proof}
\subsection{Training Objectives and Low-Noise Truncation} \label{sec:theory:interpretation}
Conditional on the sampled grid and base, the TDM distribution-matching objective corresponds directly to $D_b$. OPD velocity regression on the same $Z_T$, with the weight in~\eqref{eq:theory:weight} and the teacher in Assumption~\ref{as:theory:teacher}, corresponds directly to $O_b$. The self-consistency residual $R_b$ measures the discrepancy between the student network velocity and the velocity induced by its own forward-noised distribution. Both objectives share the student parameters, so $D_b$, $O_b$, and $R_b$ can change together during training. Theorem~\ref{thm:theory:decomposition} relates these quantities to the base KL, and identifies their exact sum when $R_b=0$.
\paragraph{Why low-noise velocity MSE.}
For a fixed interval endpoint $t_i$, the weight satisfies
\begin{align}
w_b(t)&=\frac{t_i}{t}-t_i-1+t,\label{eq:theory:weight_expansion}\\
w_b'(t)&=1-\frac{t_i}{t^2}<0,\qquad b<t<t_i\le1,\label{eq:theory:weight_derivative}\\
w_b(t_i)&=0,\qquad
w_b(t)\sim\frac{t_i}{t}\quad (b=0,\ t\downarrow0). \label{eq:theory:weight_endpoints}
\end{align}
The strict inequality follows from $t^2<t_i^2\le t_i$. Thus the weight decreases toward the noisier end of each interval, while the factor $(1-t)/t$ also emphasizes small absolute noise levels. In~\eqref{eq:theory:decomposition}, $D_b$ is the KL at a random smoothing level, and the remaining dissipation is the weighted discrepancy between $v_q$ and $v_p$. Under teacher consistency,~\eqref{eq:theory:bound_eta} bounds this remainder by $(1+\eta)O_b+(1+\eta^{-1})R_b$; when $R_b=0$, it equals $O_b$. Weighted velocity MSE therefore supplies an explicit objective for this remainder under self-consistency, with larger weights at lower noise levels. This motivates low-noise velocity supervision. We set the threshold to $0.5$. Keeping only times below that threshold approximates the full $O_b$ by its first term in
\begin{equation}
\begin{aligned}
O_b
&=\E\big[w_b(T)\|v_\theta-v_\phi\|^2\mathbf 1_{\{T\le0.5\}}\big]\\
&\quad+\E\big[w_b(T)\|v_\theta-v_\phi\|^2\mathbf 1_{\{T>0.5\}}\big].
\end{aligned} \label{eq:theory:threshold_decomposition}
\end{equation}
where all velocities are evaluated at $(Z_T,T)$.
\paragraph{Numerical truncation.}
If numerical truncation removes the interval $(b,\delta)$ for some $b<\delta<t_i$, then~\eqref{eq:theory:integrated_base} gives
\begin{equation}
\KL(q_b\|p_b)=\KL(q_\delta\|p_\delta)
+\int_b^\delta\frac{1-t}{t}m(t)\,\dd t. \label{eq:theory:truncation}
\end{equation}
The integral is the low-noise dissipation residual omitted by truncation. A decomposition based on the retained noise interval must include this residual and use the sampling normalization for that interval.

\subsection{Relation between the theorem and the implemented objective} \label{sec:theory:gap}
Theorem~\ref{thm:theory:decomposition} is stated in terms of the velocity fields $v_q$ and $v_p$ that the two forward-smoothed laws induce, while the training objectives use the network velocities $v_\theta$ and $v_\phi$. Replacing $v_q-v_p$ with $v_\theta-v_\phi$ requires two conditions. Teacher consistency (Assumption~\ref{as:theory:teacher}) identifies $v_\phi$ with $v_p$, and student self-consistency identifies $v_\theta$ with $v_q$. Without the second condition the bound in~\eqref{eq:theory:bound_eta} keeps the residual $R_b$. The specialist targets are classifier-free guided, so applying teacher consistency to them assumes that the guided network field corresponds to the induced field of the forward-smoothed teacher law.

The implemented objective in Sec.~\ref{sec:method:vmse} also departs from the weighted discrepancy in two ways. Eq.~\ref{eq:vmse_loss} averages the squared velocity difference elementwise, so it omits the continuous weight $w_b$ and the discrete weight $\omega_i$. Eq.~\ref{eq:vmse_gate} then applies the threshold of~\eqref{eq:theory:threshold_decomposition} to the mean sampled time $\bar\tau$ of the mini-batch, so the regression is active on low-noise batches. The threshold $0.5$ and the loss weight $w_{\mathrm v}$ are therefore empirical design choices. For fixed $b>0$, $w_b$ is bounded on $b<t<t_i$ and increases as $t$ decreases toward $b$, and it diverges as $t\to0$ only in the limit $b\to0$. Theorem~\ref{thm:theory:decomposition} accordingly serves as the motivation for low-noise velocity regression, and Sec.~\ref{sec:exp:ablation} reports the measured effect of the implemented objective.

\section{Training algorithm}
\label{sec:app:algorithm}

\begin{algorithm}[H]
\caption{Main-configuration training procedure for \ours{}.}
\label{alg:stepx_opd}
\fontsize{8}{8.8}\selectfont
\SetAlgoNlRelativeSize{0}

\KwIn{
learning rate $\lambda$,
training sources $\{\mathcal C_d\}$ with routing $r$,
grid size $N$,
guidance scale $w$,
velocity-loss weight $w_{\mathrm v}$,
teacher-warmup steps $s_{\mathrm w}$,
velocity-loss start step $s_{\mathrm v}$ and batch-mean threshold $\tau_{\mathrm v}$,
student $v_\theta^{\mathrm{stu}}$,
fake velocity $v_\psi^{\mathrm{fake}}$,
frozen initial student $v_{\theta_0}^{\mathrm{stu}}$,
frozen specialists $\{v_m^{\mathrm{tea}}\}_{m=1}^{M}$
}

\KwOut{optimized student $v_\theta^{\mathrm{stu}}$}

\For{$s\gets0$ \KwTo $S-1$}{

  Sample source $d$ and mini-batch
  $\mathcal B_d=\{c_j\}_{j=1}^{B}$ from $\mathcal C_d$,
  $\mathbf c\gets(c_j)_{j=1}^{B}$\;

  \If{$s<s_{\mathrm w}$}{
    $v^{\mathrm{tgt}}\gets v_{\theta_0}^{\mathrm{stu}}$\;
  }
  \Else{
    $v^{\mathrm{tgt}}\gets v_{r(d)}^{\mathrm{tea}}$
    \tcp*[r]{routed specialist, Sec.~\ref{sec:method:routing}}
  }

  Redraw the rollout grid
  $\boldsymbol\sigma$ with $1=\sigma_0>\cdots>\sigma_N=0$
        \tcp*[r]{random grid, Sec.~\ref{sec:method:points}}

  Sample noise $\bar z_{\sigma_0}$ from standard normal distribution\;

  $\{\bar z_{\sigma_n}\}_{n=0}^{N}
  \gets
  \operatorname{Rollout}^{\mathrm{sg}}
  (v_\theta^{\mathrm{stu}},\mathbf c;
  \boldsymbol\sigma,\bar z_{\sigma_0})$\;

  Sample $z_{\mathrm{start}}$ from
  $\{\bar z_{\sigma_n}\}_{n=0}^{N-1}$,
  with corresponding interval $(\sigma_u,\sigma_\ell)$\;

  $z_{\mathrm{end}}
  \gets
  z_{\mathrm{start}}
  +(\sigma_\ell-\sigma_u)
  v_\theta^{\mathrm{stu}}
  (z_{\mathrm{start}},\sigma_u,\mathbf c)$
  \tcp*[r]{Eq.~\ref{eq:euler_step}}

Sample base timestep $\tau_{\mathrm{base}}$ from $\sigma_\ell$ to $\sigma_u$\;

$x_{\mathrm{base}}
\gets
\operatorname{Interpolate}
(z_{\mathrm{start}},
\operatorname{sg}[z_{\mathrm{end}}],
\tau_{\mathrm{base}})$
\tcp*[r]{Eq.~\ref{eq:xbase}}

Sample timestep $\tau$ from $\tau_{\mathrm{base}}$ to $\sigma_u$\;

$z_\tau
\gets
\operatorname{ReNoise}
(x_{\mathrm{base}},\tau_{\mathrm{base}},\tau)$
\tcp*[r]{Eq.~\ref{eq:renoise}}

  $\psi
  \gets
  \operatorname{UpdateFake}
  (\psi;z_\tau,\tau,\mathbf c)$
  \tcp*[r]{default TDM fake update}
  $\tilde v^{\mathrm{tgt}}
  \gets
  v_{\mathrm{cfg}}^{\mathrm{tgt}}
  (z_\tau,\tau,\mathbf c)$
  \tcp*[r]{guidance scale $w$}

  $\Delta_\tau
  \gets
  \tilde v^{\mathrm{tgt}}
  -
  v_\psi^{\mathrm{fake}}
  (z_\tau,\tau,\mathbf c)$\;

  $\hat z
  \gets
  \operatorname{sg}[z_{\mathrm{end}}-\tau\Delta_\tau]$,
  \quad
  $\mathcal L_{\mathrm{TDM}}
  \gets
  \rho(z_{\mathrm{end}}-\hat z)$
  \tcp*[r]{pseudo-Huber $\rho$}
  
  $\mathcal L_{\mathrm v}\gets0$\;

  $\bar\tau
  \gets
  B^{-1}\sum_{j=1}^{B}\tau_j$\;

  \If(\tcp*[f]{Eq.~\ref{eq:vmse_gate}}){$s\ge s_{\mathrm v}$ \textbf{and} $\bar\tau\le\tau_{\mathrm v}$}{
    $\mathcal L_{\mathrm v}
    \gets
    \operatorname{mean}\!\left[
    \left(
    v_\theta^{\mathrm{stu}}
    (z_\tau,\tau,\mathbf c)
    -
    \operatorname{sg}
    [\tilde v^{\mathrm{tgt}}]
    \right)^{\odot2}
    \right]$
      \tcp*[r]{Eq.~\ref{eq:vmse_loss}}
  }

  $\theta
  \gets
  \theta-\lambda\nabla_\theta
  \left(
  \mathcal L_{\mathrm{TDM}}
  +w_{\mathrm v}\mathcal L_{\mathrm v}
  \right)$\;
}

\end{algorithm}

\section{Implementation details}
\label{sec:app:evaluation}

\paragraph{Evaluation protocol.} For every evaluated model, we generate four images for each GenEval record, OCR prompt, and PickScore prompt. See Sec.~\ref{sec:exp:setup} for the sampling steps, guidance scale, and resolution used to generate them. GenEval's 2,212 records cover 553 unique strings. The resulting image counts are 8,848 for GenEval, 4,072 for OCR, and 8,192 for PickScore. DeQA and the five additional measures that Tab.~\ref{tab:general_rewards} reports are evaluated on the same fixed 250-image subset from each of GenEval, OCR, and PickScore, for 750 images total. GenEval uses the official detector-based scorer; OCR uses normalized target-string matching with PaddleOCR; and PickScore is reported both in its raw scale and, when forming Average, after division by 26. CLIPScore is reported after division by 30. Average is $(\text{GenEval} + \text{OCR} + \text{DeQA}/5 + \text{PickScore}/26)/4$.

\paragraph{Function evaluations.} Because SD3.5-Medium uses classifier-free guidance, each of its forty sampling steps issues two forward passes, so the reported 80 NFEs correspond to forty sampling steps. FLUX.1-dev runs with classifier-free guidance disabled, so its forty sampling steps correspond to 40 NFEs.

\paragraph{Training sources.} The three training sources follow the prompt sets and the train/evaluation split that the reward-optimization and on-policy distillation works this paper builds on use for the same three objectives \citep{liu2025flowgrpo,li2026diffusionopd}, and the frozen specialists are the adapters those works release. Evaluation therefore reads the same held-out prompts those works report on.

\paragraph{Sample sizes.} The prompt-word recurrence measurement on benchmark prompts uses the 2,048 prompts of the PickScore benchmark, with three generations per prompt from three random seeds. We partition these prompts into four groups by what a scene that faithfully renders the prompt would be expected to contain. The first group has 1,633 prompts with no text expected (4,899 images per model). The second has 128 prompts that name an object which itself bears text (384 images). The third has 193 prompts that depict a scene where text is typical (579 images). The fourth has 94 prompts that specify particular text content (282 images). Tab.~\ref{tab:recurrence_realbench} reports the first group, and Tab.~\ref{tab:recurrence_by_group} reports all four, both with PickScore after division by 26. The group used in Fig.~\ref{fig:spillover} and in Figs.~\ref{fig:attribution}, \ref{fig:spillover_examples_a} and~\ref{fig:spillover_examples_b} is a further cut of the first group. This cut also excludes prompts that name an existing entity or that describe an object able to carry text, and it leaves 789 prompts and 2,367 images per model. Both panels of Fig.~\ref{fig:spillover} draw on this cut, and the right panel compares it with the fourth group, the 94 prompts that specify particular text content. Tab.~\ref{tab:fourstep_spillover} uses two further prompt sets written for this measurement: a slogan set of 100 prompts phrased like a slogan or title, and a descriptive set of 100 prompts with nothing to write on. It draws four seeds per prompt from each set across the six four-step models it evaluates, which gives $N=400$ in both columns.

\paragraph{Initialization definitions.}
Let $W_0$ denote the base weights, $D_{\mathrm{GE}}$, $D_{\mathrm{OCR}}$, and $D_{\mathrm{Pick}}$ the three specialist increments. The merge initialization is
\[
W_{\mathrm{merge}}=W_0+\tfrac13\bigl(D_{\mathrm{GE}}+D_{\mathrm{OCR}}+D_{\mathrm{Pick}}\bigr).
\]
The no-OCR variant retains coefficient $1/2$ on each of the other two increments, $W_0+\tfrac12(D_{\mathrm{GE}}+D_{\mathrm{Pick}})$, without renormalization. The +DeQA variant is the four-source mean, $W_0+\tfrac14(D_{\mathrm{GE}}+D_{\mathrm{OCR}}+D_{\mathrm{Pick}}+D_{\mathrm{DeQA}})$. For the SFT variants, $D_{\mathrm{SFT}}$ is the step-2350 SFT increment: additive initialization is $W_0+D_{\mathrm{SFT}}+\tfrac13(D_{\mathrm{GE}}+D_{\mathrm{OCR}}+D_{\mathrm{Pick}})$, whereas the four-model mean is $W_0+\tfrac14(D_{\mathrm{SFT}}+D_{\mathrm{GE}}+D_{\mathrm{OCR}}+D_{\mathrm{Pick}})$.

\section{Additional related work}
\label{sec:app:text-rendering-related}

\paragraph{Text rendered without being asked for.}
Work that touches this behavior approaches it from several directions. On the representation side, \citet{materzynska2022disentangling} show that CLIP's image encoder holds spelling and scene content in one space. They also show that text-to-image methods guided by CLIP spell out the single word they are conditioned on, where generation is an iterative optimization through CLIP. SpectralCLIP \citep{xu2024spectralclip} names these written words textual artifacts and suppresses them by masking the frequency band of the CLIP embedding they occupy. Censored sampling \citep{yoon2023censored} treats the behavior as something to remove at sampling time. Stable Diffusion, asked for a photo of a human, sometimes produces prominent embedded text or shows text in place of any human figure, and about 23.7\% of the samples drawn from that one hand-written prompt are judged malign. TangPoetryBench \citep{hu2026tangpoetrybench} scores text under prompts that instruct the model to include no text. It gives text integrity a dimension of its own that separates leakage of the prompt's own text from garbled characters, and it counts 119 of 1{,}280 images as carrying a text issue. On the reinforcement learning side, DDPO \citep{black2024ddpo} reports a policy trained against a vision-language reward printing the string \texttt{sixx ttutttas} over an image of eight turtles, as one qualitative case from a single-reward run. RLRF \citep{rodriguez2025rlrf} lists a text-in-image hack in which the policy emits the prompt string as an SVG text primitive instead of drawing the described scene. \citet{hong2026rewardhacking} organize the failure modes of text-to-image reinforcement learning around artifact-prone generation and train an artifact reward model as a regularizer for the reward models in use. Rendered text is not among the modes that catalogue lists. Flow-OPD \citep{fang2026flowopd} attributes degraded aesthetics under aggressive optimization of text rendering to reward hacking, citing Flow-GRPO \citep{liu2025flowgrpo}, and the evidence it reports is a drop in GenEval. We report this behavior as a count taken under one protocol on benchmark prompts that neither ask for rendered text nor forbid it, on the same backbone before and after several rewards are integrated into it.

\section{Introduction to unintended text rendering}
\label{sec:app:text-rendering}

\subsection{Prompt-word recurrence}
\label{sec:app:recurrence}

\begin{figure}[t]
\centering
\includegraphics[width=\linewidth]{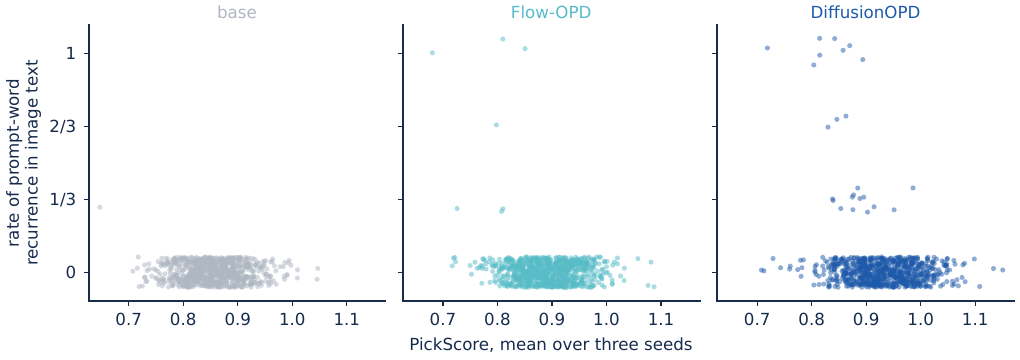}
\caption{Prompt-word recurrence against PickScore, resolved to individual prompts. Each point is one prompt on one model: the vertical coordinate is the fraction of that prompt's three seeds whose detected on-image text reproduces a word from the prompt, and the horizontal coordinate is that prompt's PickScore averaged over the same three seeds. Panels and colours are the three SD3.5-Medium models of Fig.~\ref{fig:spillover}, on the prompt group that figure draws from. Points carry a small vertical jitter so that points sharing a vertical value are visible as separate marks.}
\label{fig:fourstep_corr}
\end{figure}

\paragraph{Prompt-word recurrence on prompts with no text called for.}
Tab.~\ref{tab:recurrence_realbench} reports what eight models write into the image on prompts that ask for no rendered text. Seven of them are sampled at forty steps and \ours{} at its four-step working point. The prompts are the 2,048 that ship with the PickScore benchmark, stratified by a rule that reads the prompt text alone. The stratum reported here is the one where no text is expected, whose described scenes hold no surface able to carry text. We read every generation with the same OCR service that supplies the training reward. A generation counts as prompt-word recurrence when its detected text contains an adjacent pair of content words from the prompt, or any six-character contiguous substring of the prompt. The untrained SD3.5-Medium base renders text in 166 generations and reproduces a prompt word in 21 of them. SD3.5-Medium DiffusionOPD renders text in 699 and reproduces a prompt word in 218. Fig.~\ref{fig:spillover} draws from a further cut of this stratum that also excludes prompts that name an existing entity or that describe an object able to carry text. Tab.~\ref{tab:recurrence_by_group} gives the same two counts on all four strata, and Fig.~\ref{fig:fourstep_corr} resolves the measurement to individual prompts.
\begin{table}[t]
\centering
\footnotesize
\caption{Comparison of text rendered, the count of generations with any detected on-image text, and prompt-word recurrence, the count among those whose detected text reproduces a word from the prompt, across models, on PickScore benchmark prompts that do not call for rendered text, together with each model's PickScore.}
\label{tab:recurrence_realbench}
\begin{tabular}{lrrr}
\toprule
Model & Text rendered & Prompt-word recurrence & PickScore $\uparrow$ \\
\midrule
SD3.5-Medium base            & 166 & 21  & 0.838 \\
SD3.5-Medium SFT (GenEval)   & 283 & 40  & 0.838 \\
SD3.5-Medium DiffusionOPD    & 699 & 218 & 0.912 \\
SD3.5-Medium Flow-OPD        & 238 & 66  & 0.872 \\
\midrule
FLUX.1-dev base               & 177 & 8  & 0.844 \\
FLUX.1-dev DiffusionOPD       & 266 & 44 & 0.907 \\
FLUX.1-dev Flow-OPD           & 276 & 54 & 0.904 \\
\midrule
\ours{} (four-step)          & 304 & 94  & 0.885 \\
\bottomrule
\end{tabular}
\end{table}

\begin{figure}[t]
\centering
\includegraphics[width=\linewidth]{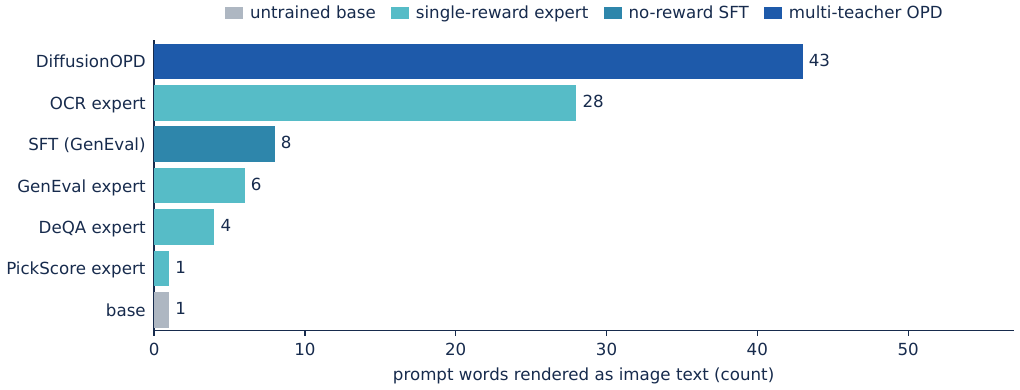}
\caption{Count of generated images whose detected text reproduces a word drawn from the prompt, for seven models on the prompt group that calls for no rendered text. Colour marks the kind of post-training each model received.}
\label{fig:attribution}
\end{figure}

\paragraph{Prompt-word recurrence measurements.}
Tab.~\ref{tab:recurrence_by_group} extends Tab.~\ref{tab:recurrence_realbench} to all four prompt groups, so the counts on the group that calls for no rendered text can be read against the group whose prompts state what text should appear. Fig.~\ref{fig:attribution} reports seven models evaluated under one protocol at forty sampling steps. For each model it gives the count of generated images whose detected text reproduces a word drawn from the prompt, on the benchmark prompt group that Fig.~\ref{fig:spillover} draws from. Tab.~\ref{tab:fourstep_spillover} reports the same measurement as a fraction for six four-step models. It uses a slogan set and a descriptive set with nothing to write on, both written for this measurement, and it also lists the OCR and GenEval scores. Fig.~\ref{fig:fourstep_corr} resolves the measurement to individual prompts for the three forty-step models of Fig.~\ref{fig:spillover}. Each point places one prompt's recurrence rate over three seeds against that prompt's PickScore over the same three seeds, and PickScore is the quality measure these prompts carry.
\begin{table*}[t]
\centering
\fontsize{6}{7.5}\selectfont
\setlength{\tabcolsep}{1.2pt}
\caption{Comparison of text rendered, the count of generations with any detected on-image text, and prompt-word recurrence, the count among those whose detected text reproduces a word from the prompt, together with PickScore, across models, for four prompt groups defined by whether the described scene calls for rendered text.}
\label{tab:recurrence_by_group}
\begin{tabular}{lrrrrrrrrrrrr}
\toprule
 & \multicolumn{3}{c}{\shortstack{No text\\expected}} & \multicolumn{3}{c}{\shortstack{Text-bearing\\object named}} & \multicolumn{3}{c}{\shortstack{Scene-typical\\text}} & \multicolumn{3}{c}{\shortstack{Specified text\\content}} \\
\cmidrule(lr){2-4} \cmidrule(lr){5-7} \cmidrule(lr){8-10} \cmidrule(lr){11-13}
Model & \shortstack{Text\\rendered} & \shortstack{Prompt-word\\recurrence} & \shortstack{Pick-\\Score $\uparrow$} & \shortstack{Text\\rendered} & \shortstack{Prompt-word\\recurrence} & \shortstack{Pick-\\Score $\uparrow$} & \shortstack{Text\\rendered} & \shortstack{Prompt-word\\recurrence} & \shortstack{Pick-\\Score $\uparrow$} & \shortstack{Text\\rendered} & \shortstack{Prompt-word\\recurrence} & \shortstack{Pick-\\Score $\uparrow$} \\
\midrule
SD3.5-Medium base            & 166 & 21  & 0.838 & 149 & 54  & 0.821 & 74  & 2   & 0.842 & 234 & 126 & 0.860 \\
SD3.5-Medium SFT (GenEval)   & 283 & 40  & 0.838 & 162 & 78  & 0.820 & 103 & 12  & 0.840 & 247 & 137 & 0.872 \\
SD3.5-Medium DiffusionOPD    & 699 & 218 & 0.912 & 273 & 178 & 0.889 & 198 & 44  & 0.909 & 247 & 159 & 0.928 \\
SD3.5-Medium Flow-OPD        & 238 & 66  & 0.872 & 159 & 77  & 0.850 & 80  & 11  & 0.870 & 255 & 155 & 0.895 \\
\midrule
FLUX.1-dev base               & 177 & 8  & 0.844 & 66  & 12  & 0.823 & 84  & 3  & 0.848 & 225 & 115 & 0.871 \\
FLUX.1-dev DiffusionOPD       & 266 & 44 & 0.907 & 144 & 63  & 0.891 & 141 & 25 & 0.906 & 242 & 133 & 0.927 \\
FLUX.1-dev Flow-OPD           & 276 & 54 & 0.904 & 151 & 60  & 0.887 & 156 & 17 & 0.902 & 242 & 132 & 0.928 \\
\midrule
\ours{} (four-step)          & 304 & 94  & 0.885 & 200 & 73 & 0.858 & 74  & 17 & 0.880 & 253 & 141 & 0.895 \\
\bottomrule
\end{tabular}
\end{table*}

\begin{table}[t]
\centering
\footnotesize
\caption{Comparison of prompt-word recurrence, the fraction of generations whose detected on-image text reproduces words from the prompt, across four-step models under two prompt registers, alongside each model's OCR and GenEval scores from the full evaluation.}
\label{tab:fourstep_spillover}
\begin{tabular}{lrrrr}
\toprule
 & \multicolumn{2}{c}{Prompt-word recurrence} & & \\
\cmidrule(lr){2-3}
Model & Slogan register & \shortstack{Descriptive,\\nothing to write on} & OCR $\uparrow$ & GenEval $\uparrow$ \\
\midrule
Base + TDM                      & 0.168 & 0.000 & 0.335 & 0.653 \\

DiffusionOPD + TDM              & 0.782 & 0.000 & 0.833 & 0.838 \\
Flow-OPD + TDM                  & 0.675 & 0.000 & 0.751 & 0.875 \\
SFT + TDM      & 0.610 & 0.000 & 0.725 & 0.932 \\
\midrule
\ours, w/o precise teacher   & 0.820 & 0.000 & 0.899 & 0.944 \\
\ours, w/ precise teacher      & 0.610 & 0.000 & 0.895 & 0.945 \\
\bottomrule
\end{tabular}
\end{table}

\begin{figure*}[p]
\centering
\includegraphics[width=\textwidth]{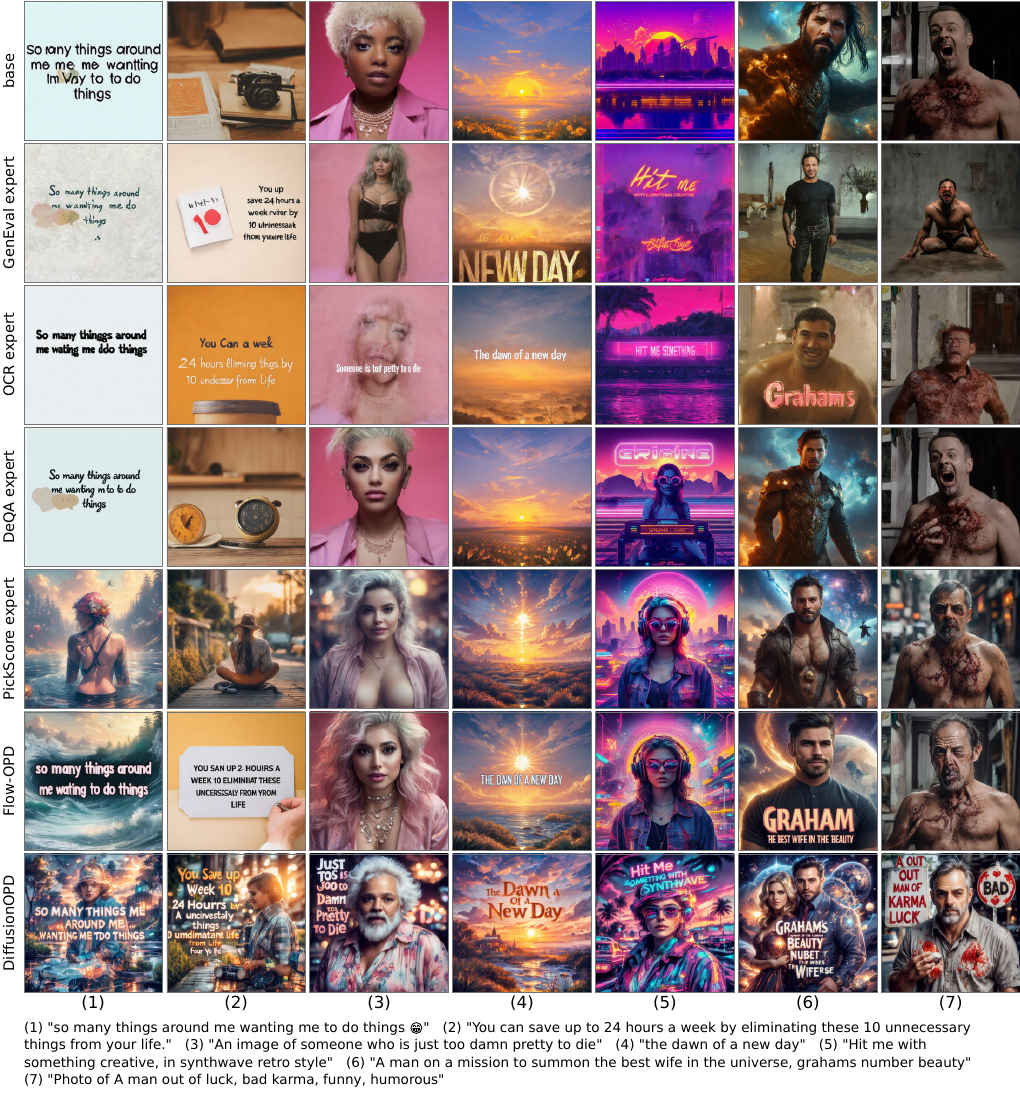}
\caption{Text seven SD3.5-Medium models render into the image, on seven of the prompts of the Fig.~\ref{fig:spillover} subset for which DiffusionOPD produces an image whose text reproduces a word from the prompt. Columns are prompts (1)--(7), printed in full below the block; rows are the untrained base, four single-reward experts, and two published multi-capability models. Fig.~\ref{fig:spillover_examples_b} continues with prompts (8)--(14).}
\label{fig:spillover_examples_a}
\end{figure*}

\begin{figure*}[p]
\centering
\includegraphics[width=\textwidth]{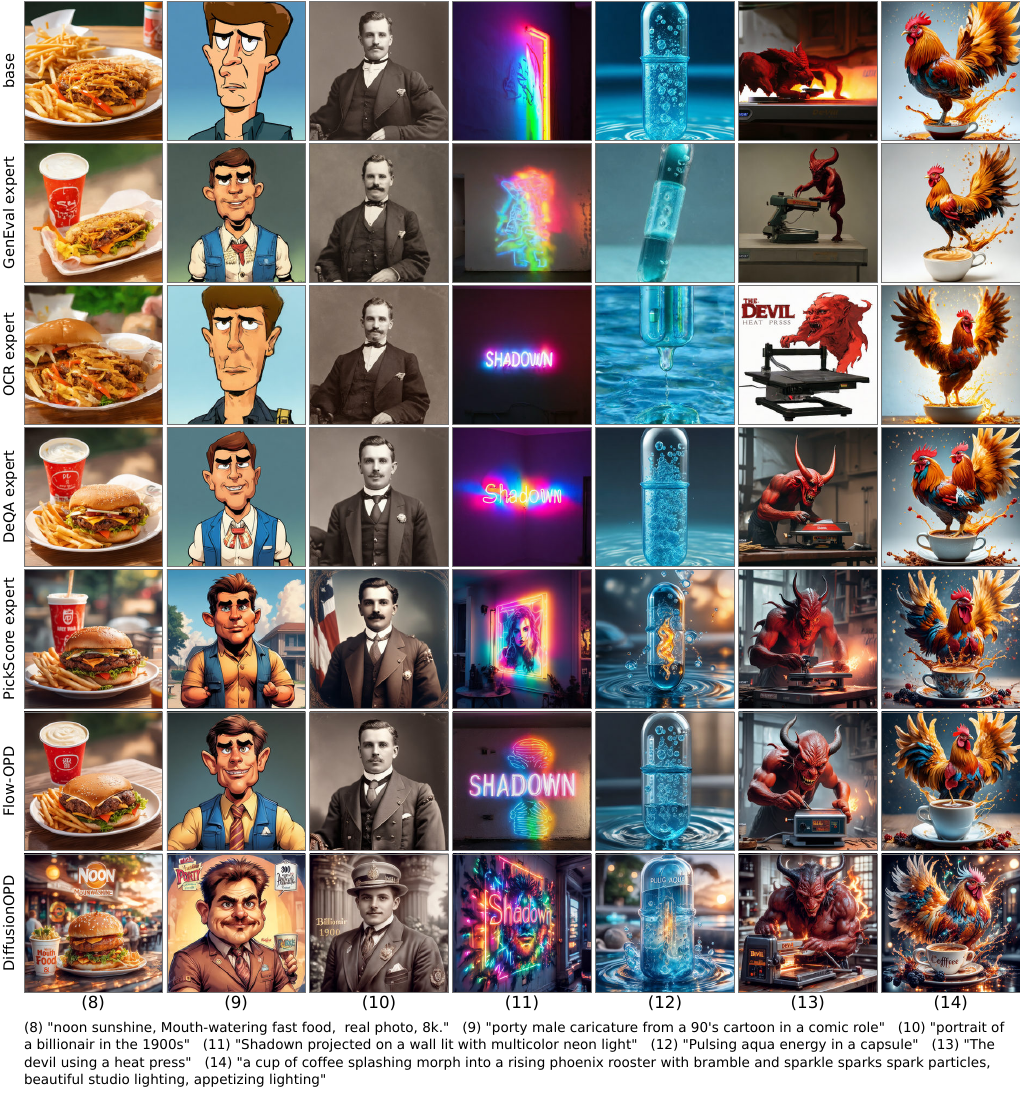}
\caption{Text seven SD3.5-Medium models render into the image, continuing Fig.~\ref{fig:spillover_examples_a} with prompts (8)--(14) of the same fourteen. Columns are those prompts, printed in full below the block; rows are the untrained base, four single-reward experts, and two published multi-capability models.}
\label{fig:spillover_examples_b}
\end{figure*}

\paragraph{Qualitative examples on the no-text-called-for subset.}
Figs.~\ref{fig:spillover_examples_a} and~\ref{fig:spillover_examples_b} extend the left panel of Fig.~\ref{fig:spillover} in two directions. The rows grow from three models to seven by adding the four single-reward experts between the untrained base and the two published multi-capability models, which shows which reward the rendered text follows. The columns are fourteen prompts of this subset on which DiffusionOPD renders a word from the prompt, and each prompt is printed in full below its block.

\subsection{Precise teacher targets and prompt-word recurrence}
\label{sec:app:precise}

At the four-step working point, each sampling step spans a much larger state interval than a step of a full-step sampler. A single instantaneous teacher velocity then describes only one point of the local transition. The precise teacher target replaces that single evaluation with an average of guided velocities recomputed along a short rollout launched from the supervision state, so one target covers the whole transition. This component is optional, and Tab.~\ref{tab:main_results} reports \ours{} without it. This subsection first gives its construction and then reports its effect on the text these models write into images.

The construction starts from each supervision point $z_\tau$ and rolls out the active target velocity field $v^{\mathrm{tgt}}$. Set $x^{(0)}=z_\tau$ and define
\begin{equation}
t_k = \frac{K-k}{K}\,\tau, \qquad k=0,\dots,K, \label{eq:teacher_times}
\end{equation}
so that $t_K=0$. At each $k=0,\dots,K-1$, evaluate the conditional and unconditional velocities at the same current state:
\begin{equation}
v_c^{\mathrm{tgt},(k)} = v^{\mathrm{tgt}}\big(x^{(k)},t_k,c\big), \qquad v_u^{\mathrm{tgt},(k)} = v^{\mathrm{tgt}}\big(x^{(k)},t_k,\varnothing\big), \label{eq:teacher_cfg_eval}
\end{equation}
Combine the two velocities with classifier-free guidance,
\begin{equation}
v_{\mathrm{cfg}}^{\mathrm{tgt},(k)} = v_u^{\mathrm{tgt},(k)} + w\,\big(v_c^{\mathrm{tgt},(k)} - v_u^{\mathrm{tgt},(k)}\big), \label{eq:cfg_velocity}
\end{equation}
and use the explicit equal-interval Euler update
\begin{equation}
 x^{(k+1)} = x^{(k)} + (t_{k+1}-t_k) v_{\mathrm{cfg}}^{\mathrm{tgt},(k)}, \label{eq:teacher_advance}
\end{equation}
recomputing both velocities at each new state. After $K$ steps, define $\hat{x}_0=x^{(K)}$ and average the guided velocities:
\begin{equation}
\bar v^{\mathrm{tgt}} = \frac{1}{K}\sum_{k=0}^{K-1} v_{\mathrm{cfg}}^{\mathrm{tgt},(k)}, \label{eq:teacher_avg}
\end{equation}
which requires $2K$ target-model forward evaluations in total.

Because the guided velocity is recomputed before every advance, the rollout follows the current guided direction at each step, and its average satisfies
\begin{equation}
z_\tau - \tau\, \bar v^{\mathrm{tgt}} = \hat{x}_0, \label{eq:secant}
\end{equation}
where $\hat{x}_0$ is the state reached at $t=0$. Thus $\bar v^{\mathrm{tgt}}$ is a secant velocity, the average slope from $z_\tau$ to the guided trajectory's endpoint, rather than an instantaneous velocity at one point. Turning the precise teacher on makes both terms of Eq.~\ref{eq:total_loss} take $\bar v^{\mathrm{tgt}}$ in place of the instantaneous target $\tilde v^{\mathrm{tgt}}$, which costs $2K-2$ target-model forwards beyond the two that the instantaneous target uses.

\paragraph{Measured overhead.}
The precise teacher adds 4.33 seconds per step inside its 500-step window, 4.8\% of ordinary training time.

\paragraph{Effect on prompt-word recurrence.}
Tab.~\ref{tab:precise_teacher_recurrence} reports the same two counts as Tab.~\ref{tab:recurrence_realbench} and Tab.~\ref{tab:recurrence_by_group}, on the same two prompt groups, for three four-step distillation configurations each evaluated with precise teacher supervision off and on. Recurrence falls in all six comparisons, from 94 to 84 and from 20 to 19 on the configuration \ours{} uses.
\begin{table}[t]
\centering
\footnotesize
\caption{Comparison of text rendered and prompt-word recurrence with precise-teacher supervision off and on, across three distillation configurations, on the prompt group with no text expected and on its subset with no text called for. Bold marks the lower prompt-word recurrence count within each configuration line.}
\label{tab:precise_teacher_recurrence}
\setlength{\tabcolsep}{4pt}
\begin{tabular}{llcrrrr}
\toprule
 & & & \multicolumn{2}{c}{No text expected} & \multicolumn{2}{c}{No text called for} \\
\cmidrule(lr){4-5} \cmidrule(lr){6-7}
Initialization & Teachers & \shortstack{Precise\\teacher} & \shortstack{Text\\rendered} & \shortstack{Prompt-word\\recurrence} & \shortstack{Text\\rendered} & \shortstack{Prompt-word\\recurrence} \\
\midrule
\multirow{2}{*}{\shortstack[l]{SFT-augmented\\merge}}       & \multirow{2}{*}{DiffusionOPD} & Off & 304 & 94           & 96 & 20           \\
                                                             &                                & On  & 298 & \textbf{84}  & 92 & \textbf{19}  \\
\midrule
\multirow{2}{*}{\shortstack[l]{Averaged-specialist\\merge}} & \multirow{2}{*}{DiffusionOPD} & Off & 284 & 101          & 93 & 24           \\
                                                             &                                & On  & 292 & \textbf{87}  & 86 & \textbf{21}  \\
\midrule
\multirow{2}{*}{\shortstack[l]{Averaged-specialist\\merge}} & \multirow{2}{*}{Flow-OPD}     & Off & 190 & 71           & 60 & 17           \\
                                                             &                                & On  & 236 & \textbf{70}  & 71 & \textbf{15}  \\
\bottomrule
\end{tabular}
\end{table}

\section{Additional experiments}
\label{sec:app:additional}

The evaluations in this section follow the protocol in App.~\ref{sec:app:evaluation}.

\paragraph{Additional quality measures.}
Tab.~\ref{tab:general_rewards} reports, for the models of both base architectures, aligned quality and preference measures that lie outside the Average score.
\begin{table*}[t]
\centering
\scriptsize
\setlength{\tabcolsep}{3pt}
\caption{Comparison of additional aligned quality and preference measures across methods, for both base architectures evaluated.}
\label{tab:general_rewards}
\begin{tabular}{llrrrrrr}
\toprule
Base Model & Model & NFEs & CLIPScore $\uparrow$ & HPSv2.1 $\uparrow$ & Aesthetic $\uparrow$ & ImageReward $\uparrow$ & UnifiedReward $\uparrow$ \\
\midrule
 \multirow{11}{*}{SD3.5-Medium}
 & Base               & 80 & 1.000 & 0.289 & 5.552 & 1.081 & 3.701 \\
 & GenEval Teacher    & 80 & 1.020 & 0.277 & 5.541 & 1.222 & 3.783 \\
 & OCR Teacher        & 80 & 0.994 & 0.274 & 5.390 & 1.015 & 3.579 \\
 & PickScore Teacher  & 80 & 1.004 & 0.325 & 5.954 & 1.396 & 3.964 \\
 & DeQA Teacher       & 80 & 1.011 & 0.301 & 5.622 & 1.293 & 3.903 \\
 & DiffusionOPD       & 80 & 1.033 & 0.304 & 5.676 & 1.288 & 3.874 \\
 & Flow-OPD           & 80 & 1.021 & 0.313 & 5.788 & 1.389 & 3.992 \\
\cmidrule{2-8}
 & Base + TDM         & 4  & 0.983 & 0.285 & 5.567 & 0.987 & 3.524 \\
 & Flow-OPD + TDM     & 4  & 1.012 & 0.304 & 5.811 & 1.344 & 3.880 \\
 & DiffusionOPD + TDM & 4  & 1.026 & 0.307 & 5.893 & 1.263 & 3.849 \\
 \rowcolor{cyan!8}
  \cellcolor{white} & \ours{}            & 4  & 1.024 & 0.307 & 5.772 & 1.301 & 3.894 \\
\midrule
 \cellcolor{white} \multirow{10}{*}{FLUX.1-dev}
 & Base                 & 40 & 0.969 & 0.299 & 5.784 & 1.086 & 3.769 \\
 & GenEval Specialist   & 40 & 0.997 & 0.303 & 5.795 & 1.292 & 3.902 \\
 & OCR Specialist       & 40 & 0.982 & 0.298 & 5.713 & 1.138 & 3.798 \\
 & PickScore Specialist & 40 & 1.000 & 0.325 & 6.101 & 1.412 & 3.942 \\
 & Joint Flow-OPD       & 40 & 1.017 & 0.310 & 5.790 & 1.355 & 3.987 \\
 & Joint DiffusionOPD   & 40 & 1.013 & 0.310 & 5.804 & 1.332 & 3.981 \\
\cmidrule{2-8}
 & Base + TDM           & 4  & 0.986 & 0.297 & 5.679 & 1.107 & 3.515 \\
 & Flow-OPD + TDM       & 4  & 1.017 & 0.308 & 5.642 & 1.327 & 3.800 \\
 & DiffusionOPD + TDM   & 4  & 1.020 & 0.307 & 5.662 & 1.329 & 3.771 \\
  \rowcolor{cyan!8}
  \cellcolor{white}& \ours{}              & 4  & 1.021 & 0.309 & 5.794 & 1.344 & 3.923 \\
\bottomrule
\end{tabular}
\end{table*}

\paragraph{Initialization recipe and teacher family.}
Tab.~\ref{tab:init_teacher_family} reports four four-step full-benchmark evaluations that cross two initialization recipes with two multi-teacher on-policy distillation teacher families. The averaged-specialist merge pools several published single-reward LoRA adapters with uniform weight over the attention projections. The SFT-augmented merge adds a supervised fine-tuning increment to a similarly pooled combination that also covers the feed-forward and embedding modules. With Flow-OPD teachers, the averaged-specialist merge and the SFT-augmented merge reach an Average of 0.8610 and 0.8772. With DiffusionOPD teachers, they reach 0.8779 and 0.8901.
\begin{table*}[t]
\centering
\footnotesize
\setlength{\tabcolsep}{5pt}
\caption{Ablation of two initialization recipes crossed with two multi-teacher on-policy distillation teacher families, reported as full-benchmark scores.}
\label{tab:init_teacher_family}
\begin{tabular}{llrrrrr}
\toprule
Initialization & Teachers & GenEval $\uparrow$ & OCR $\uparrow$ & DeQA $\uparrow$ & PickScore $\uparrow$ & Average $\uparrow$ \\
\midrule
\multirow{2}{*}{Averaged-specialist merge} & Flow-OPD    & 0.885 & 0.874 & 4.073 & 22.61 & 0.8610 \\
 & DiffusionOPD & 0.903 & 0.896 & 4.164 & 22.89 & 0.8779 \\
  \multirow{2}{*}{SFT-augmented merge}       & Flow-OPD    & 0.932 & 0.886 & 4.077 & 22.75 & 0.8772 \\
  & DiffusionOPD & 0.944 & 0.899 & 4.170 & 22.98 & 0.8901 \\
\bottomrule
\end{tabular}
\end{table*}

\paragraph{Training-source sampling ratio.}
Tab.~\ref{tab:historical_compound_results} reports four-step scores at different sampling ratios over the three training sources, GenEval, OCR, and PickScore, grouped by student configuration. Within the merged-initialization configuration, the equal ratio reaches an Average of 0.8901 against 0.8427 at 1:2:1.
\begin{table*}[t]
\centering
\footnotesize
\setlength{\tabcolsep}{3pt}
\caption{Four-step scores at different training-source sampling ratios (GenEval:OCR:PickScore) for two student configurations. Best results in each group are in bold.}
\label{tab:historical_compound_results}
\begin{tabular}{llrrrrr}
\toprule
Student configuration & Source ratio & GenEval $\uparrow$ & OCR $\uparrow$ & DeQA $\uparrow$ & PickScore $\uparrow$ & Average $\uparrow$ \\
\midrule
\multirow{2}{*}{Fresh-LoRA student} & 1:1:1 & 0.752 & 0.5776 & 4.129 & 22.46 & 0.7548 \\
 & 1:2:1 & 0.762 & 0.683 & 4.145 & 22.42 & 0.7839 \\
\midrule

  \cellcolor{white} \multirow{2}{*}{Merged-init.\ student} & 1:1:1 & \textbf{0.944} & \textbf{0.899} & \textbf{4.170} & \textbf{22.98} & \textbf{0.8901 }\\
 & 1:2:1 & 0.869 & 0.815 & 4.095 & 22.53 & 0.8427 \\
\bottomrule
\end{tabular}
\end{table*}

\paragraph{Soup composition.}
Tab.~\ref{tab:soup_composition} compares which specialist increments are pooled into the initialization soup.
\begin{table}[t]
\centering
\footnotesize
\caption{Ablation of soup-composition variants of the initialization, reported as full-benchmark scores.}
\label{tab:soup_composition}
\begin{tabular}{lrrrrr}
\toprule
Initialization & GenEval $\uparrow$ & OCR $\uparrow$ & DeQA $\uparrow$ & PickScore $\uparrow$ & Average $\uparrow$ \\
\midrule
Merge~(3 teacher)        & 0.8709 & 0.8603 & 4.064 & 22.53 & 0.8527 \\
Merge w/o OCR & 0.8822 & 0.8253 & 4.078 & 22.62 & 0.8483 \\
Merge w DEQA   & 0.8691 & 0.8346 & 4.086 & 22.58 & 0.8474 \\
\bottomrule
\end{tabular}
\end{table}

\paragraph{Velocity-regression evaluations.}
All rows below start velocity-MSE at step 500. The per-sample inference-grid query samples one of the four inference-grid points for each sample, and does not evaluate all four points on every training step. Tab.~\ref{tab:vmse_routing} compares this query against the continuous sampled $\tau$ query under full-range support. Switching to the per-sample inference-grid query lowers Average from 0.8195 to 0.8113.
\begin{table*}[t]
\centering
\footnotesize
\setlength{\tabcolsep}{5pt}
\caption{Comparison of two velocity-regression query designs, reported as full-benchmark scores.}
\label{tab:vmse_routing}
\begin{tabular}{lrrrrr}
\toprule
vMSE query & GenEval $\uparrow$ & OCR $\uparrow$ & DeQA $\uparrow$ & PickScore $\uparrow$ & Average $\uparrow$ \\
\midrule
Continuous sampled $\tau$        & 0.8544 & 0.7496 & 4.126 & 22.07 & 0.8195 \\
Per-sample inference-grid point  & 0.8449 & 0.7123 & 4.098 & 22.57 & 0.8113 \\
\bottomrule
\end{tabular}
\end{table*}

\paragraph{Component accumulation.}
The comparisons in the rest of this section all use SD3.5-Medium students. Tab.~\ref{tab:component_accumulation} adds the components of \ours{} one at a time on top of TDM with source routing, and also reports CDM with source routing, to measure what each component contributes. All rows share the initialization, the specialist teachers, the source ratio, and the evaluation protocol.

\paragraph{Sampling steps.}
Tab.~\ref{tab:sampling_steps} evaluates \ours{} and the two sequential baselines at one, four, and eight sampling steps, to test whether the comparison of Tab.~\ref{tab:main_results} holds at other step counts.

\paragraph{Alternative training pipelines.}
A few-step multi-capability student can also be obtained by first distilling each specialist into a four-step model with TDM and then integrating the distilled specialists by on-policy distillation, or by reinforcing a four-step generator with the GenEval, OCR, and PickScore rewards in the manner of TDM-R1 \citep{luo2026tdmr1}. Tab.~\ref{tab:distilled_specialists} reports the base model and the three specialists before and after four-step TDM distillation, which is the first stage of the former pipeline, and Tab.~\ref{tab:pipeline_order} compares the complete pipelines with the sequential baselines and \ours{}.

\paragraph{Training data.}
The TDM stage of the sequential baselines is trained on JourneyDB \citep{sun2023journeydb}, the data used by TDM, while \ours{} is trained on the Flow-OPD training prompts, whose GenEval, OCR, and PickScore sources determine the routed specialist. JourneyDB prompts carry no such source, so when \ours{} is trained on JourneyDB the specialist of each step is assigned arbitrarily. Tab.~\ref{tab:training_data} reports each method under both training sets, to compare the methods when the training data is the same.

\paragraph{Training cost.}
Tab.~\ref{tab:gpu_hours} lists, for each method, the GPU-hours spent on integrating the capabilities and on distilling to few steps. Fig.~\ref{fig:teaser} plots the same values.

\paragraph{Sharpness.}
Tab.~\ref{tab:sharpness} reports the variance of the Laplacian and the high-frequency energy ratio of four-step samples, computed for each model on the first 200 prompts of the PickScore test set with four seeds per prompt. Both measures decrease as an image becomes blurrier, and the table tests whether low-noise velocity regression reduces sharpness.
\begin{table}[t]
\centering
\footnotesize
\setlength{\tabcolsep}{4pt}
\caption{Component accumulation from routed TDM to StepX-OPD, together with routed CDM, reported as four-step full-benchmark scores. Best results are in bold.}
\label{tab:component_accumulation}
\begin{tabular}{lrrrrr}
\toprule
Configuration & GenEval $\uparrow$ & OCR $\uparrow$ & DeQA $\uparrow$ & PickScore $\uparrow$ & Average $\uparrow$ \\
\midrule
CDM + source routing & \textbf{0.9463} & 0.8288 & 4.021 & 22.49 & 0.8611 \\
\midrule
TDM + source routing & 0.9263 & 0.8488 & 4.021 & 21.09 & 0.8476 \\
+ randomized grid and interpolated re-noising & 0.9434 & 0.8656 & \textbf{4.178} & 21.95 & 0.8722 \\
+ low-noise velocity regression (StepX-OPD) & 0.9442 & \textbf{0.8986} & 4.170 & \textbf{22.98} & \textbf{0.8901} \\
\bottomrule
\end{tabular}
\end{table}

\begin{table}[t]
\centering
\footnotesize
\caption{StepX-OPD and the sequential baselines at one, four, and eight sampling steps, reported as full-benchmark scores. Best results in each group are in bold.}
\label{tab:sampling_steps}
\begin{tabular}{clrrrrr}
\toprule
Steps & Model & GenEval $\uparrow$ & OCR $\uparrow$ & DeQA $\uparrow$ & PickScore $\uparrow$ & Average $\uparrow$ \\
\midrule
\multirow{3}{*}{1} & Flow-OPD + TDM     & 0.728 & 0.707 & 3.840 & 20.42 & 0.7471 \\
 & DiffusionOPD + TDM & 0.734 & 0.709 & \textbf{3.859} & 21.09 & 0.7566 \\
 & StepX-OPD          & \textbf{0.736} & \textbf{0.770} & 3.789 & \textbf{21.40} & \textbf{0.7718} \\
\midrule
\multirow{3}{*}{4} & Flow-OPD + TDM     & 0.875 & 0.751 & \textbf{4.217} & 22.60 & 0.8345 \\
 & DiffusionOPD + TDM & 0.838 & 0.833 & 3.952 & \textbf{23.41} & 0.8403 \\
 & StepX-OPD          & \textbf{0.944} & \textbf{0.899} & 4.170 & 22.98 & \textbf{0.8901} \\
\midrule
\multirow{3}{*}{8} & Flow-OPD + TDM     & 0.851 & 0.775 & 4.191 & 22.58 & 0.8332 \\
 & DiffusionOPD + TDM & 0.852 & 0.850 & 4.043 & \textbf{23.39} & 0.8526 \\
 & StepX-OPD          & \textbf{0.941} & \textbf{0.902} & \textbf{4.198} & 22.89 & \textbf{0.8907} \\
\bottomrule
\end{tabular}
\end{table}

\begin{table}[t]
\centering
\footnotesize
\caption{The base model and the three specialists before and after four-step TDM distillation, reported as full-benchmark scores.}
\label{tab:distilled_specialists}
\begin{tabular}{lrrrrrr}
\toprule
Model & NFEs & GenEval $\uparrow$ & OCR $\uparrow$ & DeQA $\uparrow$ & PickScore $\uparrow$ & Average $\uparrow$ \\
\midrule
Base                       & 80 & 0.666 & 0.529 & 4.045 & 21.79 & 0.7105 \\
Base + TDM                 & 4  & 0.653 & 0.335 & 4.002 & 21.87 & 0.6573 \\
\midrule
GenEval specialist         & 80 & 0.946 & 0.597 & 3.965 & 21.66 & 0.7923 \\
GenEval specialist + TDM   & 4  & 0.903 & 0.525 & 4.095 & 21.15 & 0.7651 \\
\midrule
OCR specialist             & 80 & 0.656 & 0.924 & 3.849 & 21.66 & 0.7957 \\
OCR specialist + TDM       & 4  & 0.634 & 0.903 & 3.971 & 21.18 & 0.7865 \\
\midrule
PickScore specialist       & 80 & 0.744 & 0.629 & 4.166 & 23.30 & 0.7756 \\
PickScore specialist + TDM & 4  & 0.726 & 0.577 & 4.135 & 23.04 & 0.7540 \\
\bottomrule
\end{tabular}
\end{table}

\begin{table}[t]
\centering
\footnotesize
\setlength{\tabcolsep}{4pt}
\caption{Alternative pipelines for obtaining a four-step multi-capability student, reported as full-benchmark scores. Best results are in bold.}
\label{tab:pipeline_order}
\begin{tabular}{lrrrrr}
\toprule
Pipeline & GenEval $\uparrow$ & OCR $\uparrow$ & DeQA $\uparrow$ & PickScore $\uparrow$ & Average $\uparrow$ \\
\midrule
Multi-reward few-step reinforcement    & 0.741 & 0.794 & \textbf{4.230} & 21.94 & 0.8063 \\
Distill specialists, then integrate    & 0.843 & 0.832 & 4.022 & 21.44 & 0.8260 \\
Integrate, then distill (Flow-OPD + TDM)     & 0.875 & 0.751 & 4.217 & 22.60 & 0.8345 \\
Integrate, then distill (DiffusionOPD + TDM) & 0.838 & 0.833 & 3.952 & \textbf{23.41} & 0.8403 \\
StepX-OPD (single stage)               & \textbf{0.944} & \textbf{0.899} & 4.170 & 22.98 & \textbf{0.8901} \\
\bottomrule
\end{tabular}
\end{table}

\begin{table}[t]
\centering
\footnotesize
\setlength{\tabcolsep}{4pt}
\caption{Four-step scores of each method when trained on JourneyDB and on the Flow-OPD training prompts, reported as full-benchmark scores.}
\label{tab:training_data}
\begin{tabular}{llrrrrr}
\toprule
Model & Training data & GenEval $\uparrow$ & OCR $\uparrow$ & DeQA $\uparrow$ & PickScore $\uparrow$ & Average $\uparrow$ \\
\midrule
\multirow{2}{*}{Flow-OPD + TDM}     & JourneyDB        & 0.875 & 0.751 & 4.217 & 22.60 & 0.8345 \\
 & Flow-OPD prompts & 0.835 & 0.742 & 4.131 & 21.78 & 0.8103 \\
\midrule
\multirow{2}{*}{DiffusionOPD + TDM} & JourneyDB        & 0.838 & 0.833 & 3.952 & 23.41 & 0.8403 \\
 & Flow-OPD prompts & 0.802 & 0.808 & 3.986 & 22.16 & 0.8147 \\
\midrule
\multirow{2}{*}{StepX-OPD}          & JourneyDB        & 0.863 & 0.828 & 4.035 & 21.37 & 0.8300 \\
 & Flow-OPD prompts & 0.944 & 0.899 & 4.170 & 22.98 & 0.8901 \\
\bottomrule
\end{tabular}
\end{table}

\begin{table}[t]
\centering
\footnotesize
\caption{Training cost in GPU-hours of the capability-integration stage and the few-step distillation stage for each method.}
\label{tab:gpu_hours}
\begin{tabular}{lrrr}
\toprule
Model & Capability integration & Few-step distillation & Total \\
\midrule
Flow-OPD + TDM     & 1600.0 & 25.9 & 1625.9 \\
DiffusionOPD + TDM & 90.1   & 25.6 & 115.7 \\
StepX-OPD          & --     & --   & 72.7 \\
\bottomrule
\end{tabular}
\end{table}

\begin{table}[t]
\centering
\footnotesize
\caption{Sharpness of four-step samples, measured by the variance of the Laplacian and the high-frequency energy ratio.}
\label{tab:sharpness}
\begin{tabular}{lrr}
\toprule
Model & Laplacian variance $\uparrow$ & High-frequency energy ratio $\uparrow$ \\
\midrule
Flow-OPD + TDM                   & 1461.3 & 0.0246 \\
DiffusionOPD + TDM               & 2117.4 & 0.0330 \\
StepX-OPD w/o velocity regression & 2438.9 & 0.0313 \\
StepX-OPD                        & 2578.9 & 0.0342 \\
\bottomrule
\end{tabular}
\end{table}

\clearpage
\begin{figure}[p]
\section{Additional qualitative results}
\label{sec:app:qualitative}
Fig.~\ref{fig:qualitative-appendix-a} through Fig.~\ref{fig:qualitative-appendix-d} extend the comparison in Fig.~\ref{fig:quali-e} with sixteen further self-written prompts, four per figure, on the same four-step models. Fig.~\ref{fig:qualitative-comparison} gives the six-model comparison on benchmark prompts. Fig.~\ref{fig:qualitative-appendix-a} and Fig.~\ref{fig:qualitative-appendix-b} show eight scenes that call for no rendered text. Fig.~\ref{fig:qualitative-appendix-c} has two prompts of that kind and two that render a short phrase on an object. Fig.~\ref{fig:qualitative-appendix-d} specifies a phrase to render in each of its four prompts.

\centering
\includegraphics[width=\textwidth]{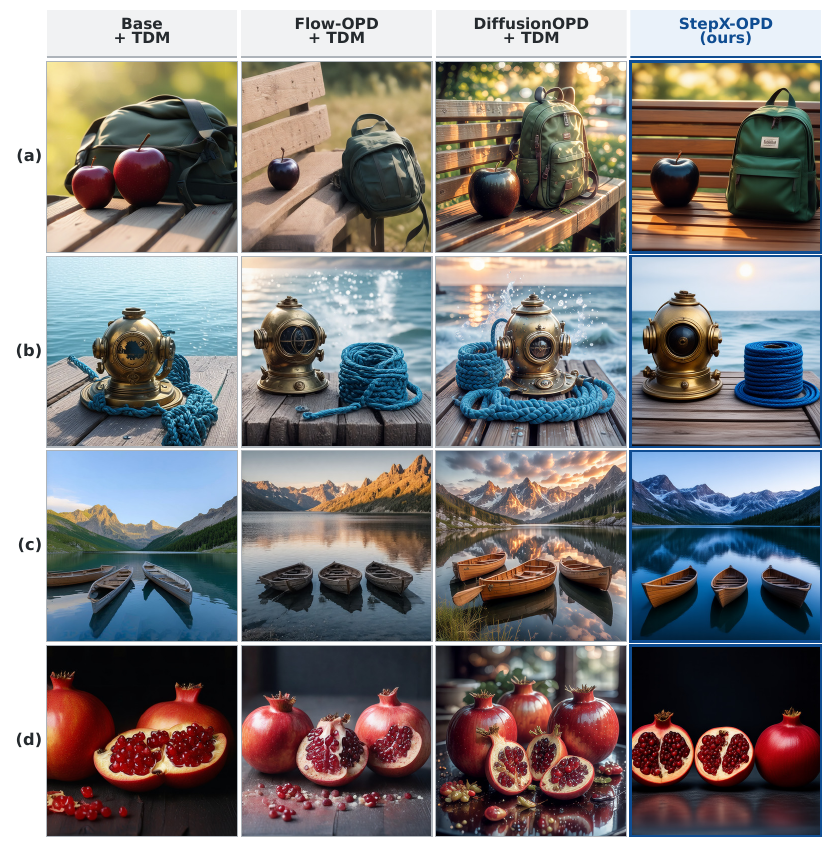}
\caption{(a) A black apple and a green backpack on a sunlit wooden bench, shallow depth of field, warm afternoon light; (b) A brass diving helmet and a coil of blue rope on a weathered dock, sea spray, low sun behind; (c) Three wooden rowing boats moored on a still alpine lake at dawn, mountains mirrored in the water; (d) Three ripe pomegranates split open on a dark table, seeds glistening, dramatic side light, baroque still life.}
\label{fig:qualitative-appendix-a}
\end{figure}
\clearpage

\begin{figure}[p]
\centering
\includegraphics[width=\textwidth]{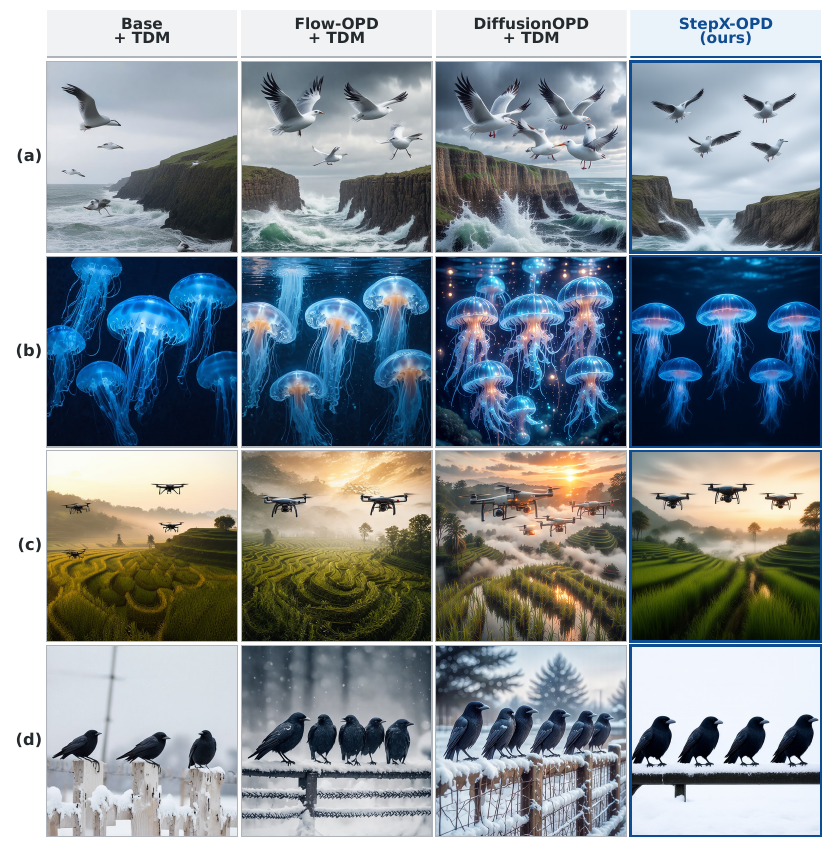}
\caption{(a) Four seagulls in flight against a stormy grey sky above cliffs, wind-torn spray below; (b) Five jellyfish drifting in dark water, translucent bells glowing blue, fine tentacles trailing; (c) Three drones hovering above a terraced rice field at sunrise, mist between the terraces; (d) Four crows lined up on a snowy fence rail, black against white, flat overcast light.}
\label{fig:qualitative-appendix-b}
\end{figure}
\clearpage

\begin{figure}[p]
\centering
\includegraphics[width=\textwidth]{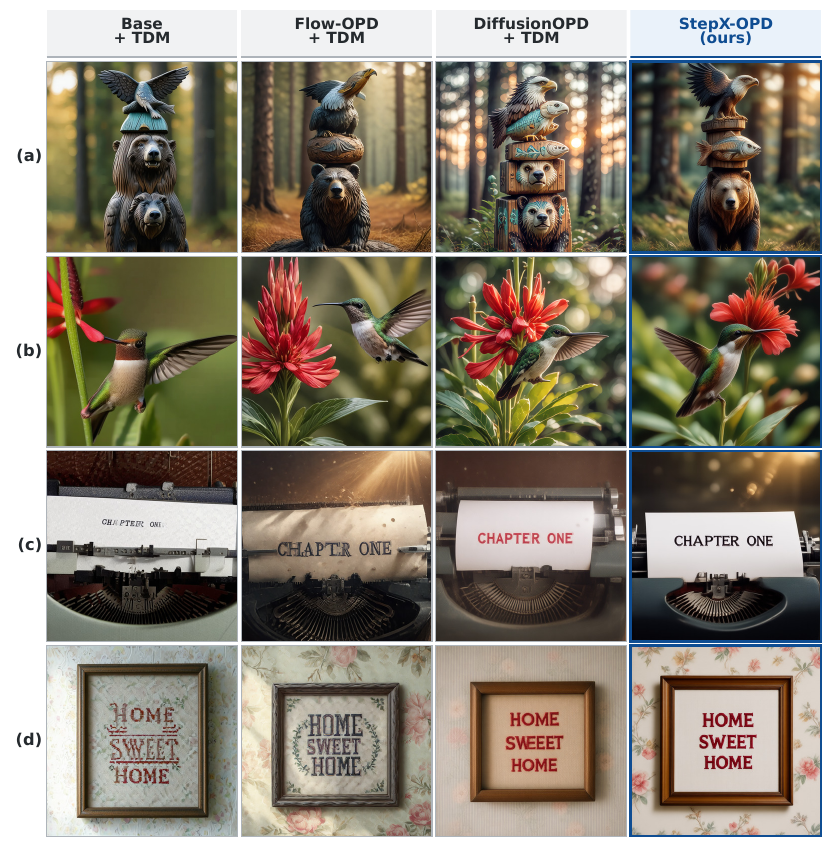}
\caption{(a) Stacked wooden animal totem with bear at the bottom, eagle in the middle and fish on top, with weathered paint, forest bokeh and warm light; (b) A hummingbird hovering in front of a red trumpet flower with blurred green foliage behind, high-speed macro; (c) An old typewriter with a sheet of paper rolled in, the typed line reading ``CHAPTER ONE'', dust motes in a slanting sunbeam; (d) A hand-stitched sampler framed on a wall reading ``HOME SWEET HOME'', late afternoon light, faded floral wallpaper behind.}
\label{fig:qualitative-appendix-c}
\end{figure}
\clearpage

\begin{figure}[p]
\centering
\includegraphics[width=\textwidth]{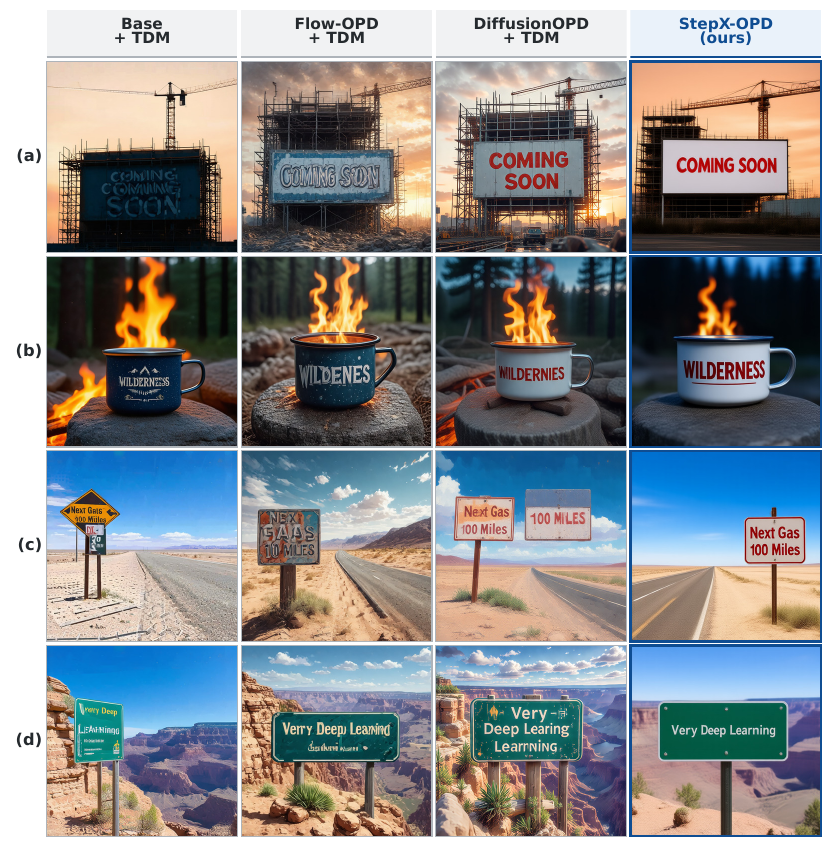}
\caption{(a) A construction site hoarding painted with ``COMING SOON'', scaffolding above, crane silhouette against an orange evening sky; (b) An enamel mug on a campfire stone printed with ``WILDERNESS'', flames glowing, pine forest dark behind, blue hour; (c) A vast, sun-baked desert landscape with a lone highway stretching into the distance. A weathered road sign stands by the side, warning ``Next Gas 100 Miles'', under a blazing sky; (d) A green sign that says ``Very Deep Learning'' and is at the edge of the Grand Canyon.}
\label{fig:qualitative-appendix-d}
\end{figure}
\clearpage

\end{document}